\documentclass{article} 
\usepackage{iclr2021_conference,times}

\usepackage{microtype}
\usepackage{graphicx}
\usepackage{subcaption}
\usepackage{arydshln}
\usepackage{booktabs} 

\usepackage{hyperref}

\usepackage[utf8]{inputenc} 
\usepackage[T1]{fontenc} 
\usepackage{hyperref} 
\usepackage{url} 
\usepackage{booktabs} 
\usepackage{amsfonts} 
\usepackage{nicefrac} 
\usepackage{microtype} 
\usepackage[utf8]{inputenc} 
\usepackage[T1]{fontenc} 
\usepackage{hyperref} 
\usepackage{url} 
\usepackage{booktabs} 
\usepackage{amsfonts} 
\usepackage{nicefrac} 
\usepackage{microtype} 
\usepackage{graphicx}
\usepackage{graphics}
\usepackage{amsthm}
\usepackage{mathrsfs}
\usepackage{amsmath}
\usepackage{amssymb}
\usepackage{mathtools}
\usepackage{wrapfig}
\usepackage{multirow}
\usepackage{soul}
\usepackage{multicol}
\usepackage[textwidth=1.2in,textsize=tiny]{todonotes}
\makeatletter
\def\BState{\State\hskip-\ALG@thistlm}
\makeatother
\newcommand{\cdotv}{\boldsymbol{\cdot}}

\newcommand{\E}{\mathbb{E}}
\newcommand{\bas}[1]{\begin{align*}#1\end{align*}}
\newcommand{{\phivmat}}{\boldsymbol{{\phi}}}
\newcommand{{\phiv}}{\boldsymbol{{\phi}}}
\newcommand{\ba}[1]{\begin{align}#1\end{align}}

\usepackage{stackengine}

\makeatletter
\newcommand{\distas}[1]{\mathbin{\overset{#1}{\kern\z@\sim}}}%

\newcommand{\ra}[1]{\renewcommand{\arraystretch}{#1}}
\newcommand{\beqs}{\vspace{0mm}\begin{eqnarray}}
\newcommand{\eeqs}{\vspace{0mm}\end{eqnarray}}
\newcommand{\barr}{\begin{array}}
\newcommand{\earr}{\end{array}}

\usepackage[ruled,vlined]{algorithm2e}
\usepackage{algorithmic}

\newcommand{\pv}[0]{{\boldsymbol{p}}}

\newcommand{\xv}{\boldsymbol{x}}

\newcommand{\zv}{\boldsymbol{z}}

\newcommand{\alphav}{\boldsymbol{\alpha}}

\newcommand{\epsilonv}{\boldsymbol{\epsilon}}

\newcommand{\etav}[0]{{\boldsymbol{\eta}}}

\newcommand{\thetav}{\boldsymbol{\theta}}

\newcommand{\piv}{\boldsymbol{\pi}}

\newcommand{\tauv}[0]{{\boldsymbol{\tau}} }

\newcommand{\varphiv}{\boldsymbol{\varphi}}

\newcommand{\given}{\,|\,}

\newcommand{\mz}[1]{\todo[color=blue!10]{mz: #1}}

\title{Contextual Dropout: An Efficient Sample-Dependent Dropout Module}

\author{Xinjie Fan$^{*,1}$, Shujian Zhang$^{*,1}$, Korawat Tanwisuth$^1$, Xiaoning Qian$^2$, Mingyuan Zhou$^1$\\
$^1$The University of Texas at Austin, $^2$Texas A\&M University, \\
\texttt{xfan@utexas.edu, szhang19@utexas.edu,}\\\texttt{korawat.tanwisuth@utexas.edu,xqian@ece.tamu.edu,}\\ \texttt{mingyuan.zhou@mccombs.utexas.edu}}

\iclrfinalcopy 
\begin{document}

\maketitle

\begin{abstract}
Dropout has been demonstrated as a simple and effective module to not only regularize the training process of deep neural networks, but also provide the uncertainty estimation for prediction. However, the quality of uncertainty estimation is highly dependent on the dropout probabilities. Most current models use the same dropout distributions across all data samples due to its simplicity. Despite the potential gains in the flexibility of modeling uncertainty, sample-dependent dropout, on the other hand, is less explored as it often encounters scalability issues or involves 
non-trivial model changes. In this paper, we propose contextual dropout with an efficient 
structural design
as a simple and scalable sample-dependent dropout module, 
which can be applied to a wide range of models 
at the expense of only slightly increased
memory and computational cost. We learn the dropout probabilities with a variational objective, compatible with both Bernoulli dropout and Gaussian dropout. We apply the contextual dropout module to various models with applications to image classification and visual question answering and demonstrate the scalability of the method with large-scale datasets, such as ImageNet and VQA~2.0. Our experimental results show that the proposed method outperforms baseline methods in terms of both accuracy and quality of uncertainty estimation.
{\let\thefootnote\relax\footnote{{$^*$ Equal contribution.
Corresponding to: \texttt{mingyuan.zhou@mccombs.utexas.edu}}}} 

\end{abstract}

\section{Introduction}
Deep neural networks (NNs) have become ubiquitous
and achieved state-of-the-art results in a wide variety of research problems 
\citep{lecun2015deep}.  
To prevent over-parameterized NNs from overfitting, we often need to appropriately regularize their training.
One way to do so is to use Bayesian NNs that treat the NN weights as random variables and regularize them with appropriate prior distributions \citep{mackay1992practical,neal2012bayesian}.
More importantly, we can obtain the model's confidence on its predictions
by evaluating the consistency between the predictions that are conditioned on different posterior samples of the NN weights.
However, despite significant recent efforts in developing various types of approximate inference for Bayesian NNs \citep{graves2011practical,welling2011bayesian,li2016preconditioned,
blundell2015weight,louizos2017multiplicative,shi2018kernel}, 
the large number of NN weights makes it difficult to 
scale to real-world applications. 


Dropout has been demonstrated as another effective regularization strategy, which can be viewed as imposing a distribution over the NN weights  \citep{gal2016dropout}.
Relating dropout to Bayesian inference 
provides a much simpler and more efficient way than using vanilla Bayesian NNs to provide uncertainty estimation 
\citep{gal2016dropout}, as there is no more need to explicitly instantiate multiple sets of NN weights. 
For example, Bernoulli dropout randomly shuts down neurons during training 
\citep{hinton2012improving,srivastava2014dropout}. 
Gaussian dropout 
multiplies the neurons with independent, and identically distributed ($iid$) Gaussian random variables drawn from $\mathcal{N}(1,\alpha)$, where the variance $\alpha$ is a tuning parameter \citep{srivastava2014dropout}. Variational dropout generalizes Gaussian dropout by 
reformulating it under a Bayesian setting and allowing $\alpha$ 
to be learned under a variational objective \citep{kingma2015variational,molchanov2017variational}. 

However, the quality of 
uncertainty estimation 
depends heavily on the dropout probabilities \citep{gal2017concrete}. To avoid grid-search over the dropout probabilities, \citet{gal2017concrete} and \citet{ boluki2020learnable} propose to automatically learn the dropout probabilities, which not only leads to a faster experiment cycle but also enables the model to have different dropout probabilities for each layer, bringing greater flexibility into uncertainty modeling. 
But, these methods still impose the restrictive assumption that dropout probabilities are global parameters shared across all data samples. 
By contrast, we consider parameterizing dropout probabilities 
as a function of input covariates, 
treating them as data-dependent local variables.
Applying covariate-dependent dropouts 
allows different data 
to have different distributions over the NN weights.  
 This generalization 
has the potential to 
greatly enhance the expressiveness of a Bayesian NN. 
However, learning covariate-dependent dropout rates is challenging. \citet{ba2013adaptive} propose \textit{standout}, where a binary belief network is laid over the original network, and develop a heuristic approximation to optimize free energy. But, as pointed out by \citet{gal2017concrete}, 
it is not scalable due to its need to significantly increase the model size. 

In this paper, we propose a simple and scalable
contextual dropout module, whose dropout rates depend on the covariates $\xv$, as a new approximate Bayesian inference method for NNs.
With a novel design that reuses the main
network to define how the covariate-dependent dropout rates are produced, 
it boosts the performance while only slightly increases the memory and computational cost. 
Our method greatly enhances the flexibility of modeling, maintains the inherent advantages of dropout over conventional Bayesian NNs, and is generally simple to implement and scalable to the large-scale applications.
We plug the  
contextual dropout module into various types of NN layers, including fully connected, convolutional, and attention layers. On a variety of supervised learning tasks, contextual dropout achieves good performance in terms of accuracy and quality of uncertainty estimation.

\section{Contextual dropout}

We introduce an efficient solution for data-dependent dropout: (1) treat the dropout probabilities as sample-dependent local random variables, (2) propose an efficient parameterization of dropout probabilities by sharing parameters between the encoder and decoder, and (3) learn the dropout distribution with a variational objective.

\subsection{Background on dropout modules}
Consider a supervised learning problem with training data $\mathcal{D}:=\{\xv_i, y_i\}_{i=1}^N$, where we model the conditional probability $p_{\thetav}(y_i\given\xv_i)$ using a NN parameterized by~$\thetav$. 
Applying dropout to a NN often means element-wisely reweighing each layer 
%
with a data-specific Bernoulli/Gaussian distributed random mask $\zv_i$, which are $iid$ drawn from a prior $p_{\etav}(\zv)$ parameterized by $\etav$ \citep{hinton2012improving,srivastava2014dropout}. 
This 
implies dropout training can be viewed as approximate Bayesian inference 
\citep{gal2016dropout}. More specifically, one may 
view the learning objective of a supervised learning model with dropout as a log-marginal-likelihood: 
$
\log \int \prod_{i=1}^N p(y_i\given \xv_i, \zv)p(\zv)d \zv$. To maximize this often intractable log-marginal, it is common to resort to variational inference \citep{hoffman2013stochastic,
blei2017variational} that introduces a variational distribution $q({\zv})$ on the random mask $\zv$ and 
optimizes an evidence lower bound (ELBO):  
\ba{\label{eq:ELBO_regular}\resizebox{0.94\hsize}{!}{$\mathcal{L}(\mathcal D)=\textstyle
\E_{q(\zv)}\left[  \log\frac{\prod_{i=1}^N p_{\thetav}(y_i\given \xv_i, \zv)p_{\etav}(\zv)}{q(\zv)}\right] 
= \big(\sum_{i=1}^N\E_{\zv_i\sim q(\zv)}\left[  \log p_{\thetav}(y_i\given \xv_i, \zv_i)\right]\big)-\mbox{KL}(q(\zv)||p_\eta(\zv)),$}
}
where ${\small\mbox{KL}(q(\zv)||p_{\etav}(\zv))=\E_{q(\zv)}[\log {q(\zv)}-\log{p(\zv)}]}$ is a Kullback--Leibler (KL) divergence based regularization term. Whether the KL term is explicitly imposed is a key distinction between regular dropout \citep{hinton2012improving,srivastava2014dropout} and their Bayesian generalizations \citep{gal2016dropout,gal2017concrete,kingma2015variational,molchanov2017variational,boluki2020learnable}.

\subsection{ 
Covariate-dependent weight uncertainty
}





In regular dropout, as shown in \eqref{eq:ELBO_regular}, while we make the dropout masks data specific during optimization, we keep their distributions the same.
This implies that while the NN weights can vary from data to data, their distribution is kept data invariant. In this paper, we propose \textit{contextual dropout}, in which the distributions of dropout masks $\zv_i$ depend on covariates $\xv_i$ for each sample $(\xv_i,y_i)$. Specifically, 
we define the variational distribution as $q_{\phiv}(\zv_i\given \xv_i)$, where $\phiv$ denotes its NN parameters. In the framework of amortized variational Bayes \citep{kingma2013auto, rezende2014stochastic}, we  can view $q_{\phivmat}$ as an inference network (encoder) trying to approximate the posterior  $p(\zv_i\given y_i, \xv_i)\propto p(y_i\given \xv_i,\zv_i)p(\zv_i)$. Note as 
we have no access to $y_i$ during testing, 
we parameterize our encoder in a way that it  
depends on $\xv_i$ but not $y_i$.  
From the optimization point of view, what we propose corresponds to the ELBO of $\log\prod_{i=1}^N  \int  p(y_i\given \xv_i,\zv_i) p(\zv_i) d\zv_i $ given $q_{\phivmat}(\zv_i \given \xv_i)$ as the encoder, 
which can be expressed as 
\ba{\small
\textstyle
\mathcal{L}(\mathcal{D}) = \sum_{i=1}^N\mathcal{L}(\xv_i,y_i),~\mathcal{L}(\xv_i,y_i) =
\E_{\zv_i\sim q_{\phivmat}(\cdotv \given \xv_i)}[\log p_{\thetav}(y_i\given\xv_i,\zv_i)] 
-\text{KL}(q_{\phivmat}(\zv_i\given\xv_i) || p_{\etav}(\zv_i)). 
\label{eq:ELBO_contextual}
}
This ELBO differs from that of regular dropout in \eqref{eq:ELBO_regular} in that 
the dropout distributions for $\zv_i$ are now parameterized by $\xv_i$ 
and 
a single KL regularization term is replaced with the aggregation of $N$ data-dependent KL terms. Unlike conventional Bayesian NNs, as $\zv_i$ is now a local random variable, the impact of the KL terms will not diminish as $N$ increases, and from the viewpoint of uncertainty quantification, contextual dropout relies only on aleatoric uncertainty to model its uncertainty on $y_i$ given $\xv_i$. Like conventional BNNs, we may add epistemic uncertainty by imposing a prior distribution on $\thetav$ and/or $\phiv$, and infer their posterior given $\mathcal{D}$. As contextual dropout with a point estimate on both $\thetav$ and $\phiv$ is already achieving state-of-the-art performance, we leave that extension for future research.
In what follows, we 
omit the data index $i$ for simplification 
and formally define its model structure.

{\bf Cross-layer dependence: } For a NN with $L$ layers, we denote $\zv=\{\zv^1,\ldots,\zv^L\}$, with $\zv^l$ representing the dropout masks at layer $l$.
As we expect 
$\zv^l$ to be dependent on the dropout masks in previous layers $\{\zv^j\}_{j<l}$, we 
introduce an autoregressive distribution as 
$q_{\phivmat}(\zv\given\xv) = \prod _{l=1}^L q_{\phivmat} (\zv^{l}\given \xv^{l-1})$, where $\xv^{l-1}$, the output of layer $l-1$,  is a function of $\{ \zv^{1},\ldots,\zv^{l-1}, \xv\}$.

\begin{wrapfigure}[9]{r}{0.5\textwidth} 
\label{fig:cap}
\vspace{-0.1in}
\centering
\includegraphics[width=0.5\textwidth]{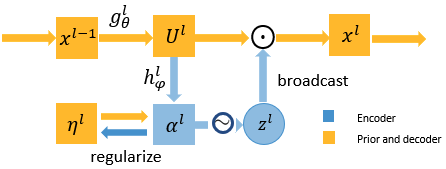}\vspace{-0.1in}
 \caption{\small 
 {A contextual dropout module.}}
 \label{fig:flow0}
\end{wrapfigure}

{\bf Parameter sharing between encoder and decoder: } 
We aim to build an encoder by modeling $q_{\phivmat} (\zv^{l}\given \xv^{l-1})$, 
where $\xv$ may come from complex and highly structured data such as images and natural languages. Thus, extracting useful features from $\xv$ to learn the encoder distribution $q_{\phivmat}$ itself becomes a problem as challenging as the original one, $i.e.$, extracting discriminative features from $\xv$ to predict $y$. 
As intermediate layers in the decoder network $p_{\thetav}$ are already learning useful features from the input, we choose to reuse them in the encoder, instead of extracting the features from scratch. If we denote layer $l$ of the decoder network by $g_{\thetav}^l$, then the output of layer~$l$, given its input $\xv^{l-1}$, 
would be $\mathbf{U}^l = g_{\thetav}^{l}(\xv^{l-1})$.
Considering this as a learned feature for $\xv$, 
as illustrated in Figure~\ref{fig:flow0}, we build the encoder on this output as 
$\alphav^l = h_{\varphiv}^l (\mathbf{U}^l)$, draw $\zv^l$ conditioning on $\alphav^l$, and element-wisely multiply $\zv^l$ with $\mathbf{U}^l$ (with broadcast if needed) to produce the output of layer $l$ as 
$\xv^l$. In this way, we use $\{\thetav, \varphiv\}$ to parameterize the encoder, 
which reuses
parameters $\thetav$ of the decoder.
To produce the dropout rates of the encoder, 
we only need extra parameters $\varphiv$, the added memory and computational cost of which are often insignificant in comparison to these of the decoder.
\subsection{Efficient parameterization of contextual dropout module} 

Denote the output of layer $l$ by a multidimensional array (tensor) $\mathbf{U}^l=g_{\thetav}^{l}(\xv^{l-1})\in \mathbb{R}^{C_1^l \times\ldots\times C_{D^l}^l}$, where $D^l$ denotes the number of the dimensions of $\mathbf{U}^l$ and $C_d^l$ denotes the number of elements along dimension $d\in\{1,\ldots,D^l\}$.
For efficiency, 
the output shape of $h_{\varphiv}^l$ is not matched to the shape of $\mathbf{U}^l$. Instead, we make it smaller and broadcast the contextual dropout masks $\zv^{l}$ across the 
dimensions of $\mathbf{U}^l$ 
\citep{tompson2015efficient}. 
Specifically, we parameterize dropout logits $\alphav^l$ of the variational distribution to have 
 $C_{d}^l$ elements, where $d\in \{1, ...., D^l\}$ is a specified dimension 
 of $\mathbf{U}^l$. We sample 
 $\zv^{l}$ from the encoder and
broadcast them across all but dimension $d$ of $\mathbf{U}^l$. We sample $\zv^{l}\sim\text{Ber} (\sigma(\alphav^l))$ under contextual Bernoulli dropout, and follow \citet{srivastava2014dropout} to use $\zv^{l}\sim N(1, \sigma(\alphav^{l})/(1-\sigma(\alphav^{l})))$ for contextual Gaussian dropout.
To obtain  $\alphav^l\in \mathbb{R}^{C_{d}^l}$, we first take the average pooling of $\mathbf{U}^l$ across all but dimension $d$, with the output denoted as $F_{\text{avepool}, d}(\mathbf{U}^l)$, 
and then apply two fully-connected layers $\Phi_1^l$ and $\Phi_2^l$ connected by $F_\text{NL}$, a 
(Leaky) ReLU based nonlinear activation function, as
\ba{\alphav^l =h_{\varphiv}^l (\mathbf{U}^l)= \Phi_2^l( F_\text{NL}(\Phi_1^l(F_{\text{avepool}, d}(\mathbf{U}^l)))),}
where $\Phi_1^l$ is a linear transformation mapping from $\mathbb{R}^{C^l_d}$ to $\mathbb{R}^{C^l_d/\gamma}$, while $\Phi_2^l$ is from $\mathbb{R}^{C^l_d/\gamma}$ back to $\mathbb{R}^{C^l_d}$, with $\gamma$ being a reduction ratio controlling the complexity of $h_{\varphiv}^l$. Below we describe how to apply contextual dropout to three representative types of NN layers.

\textit{Contextual dropout module for fully-connected layers\footnote{{Note that full-connected layers can be applied to multi-dimensional tensor as long as we specify the dimension along which the summation operation is conducted \citep{abadi2016tensorflow}.}}:} 
If layer $l$ 
is a fully-connected layer and $\mathbf{U}^l \in \mathbb{R}^{C_1^l \times\cdots\times C_{D^l}^l}$, we set $\alphav^l\in \mathbb{R}^{C^l_{D^l}}$, where 
$D^l$ is the dimension that the linear transformation is applied to. Note, if $\mathbf{U}^l \in \mathbb{R}^{C^l_1}$, then $\alphav^l\in \mathbb{R}^{C^l_1}$, and $F_{\text{avepool}, 1}$ is an identity map, so
$\alphav^l = \Phi_2^l (F_\text{NL}(\Phi_1^l(\mathbf{U}^l))).$ 

 \begin{figure}[t]
 \centering
 \includegraphics[width=1.0\textwidth]{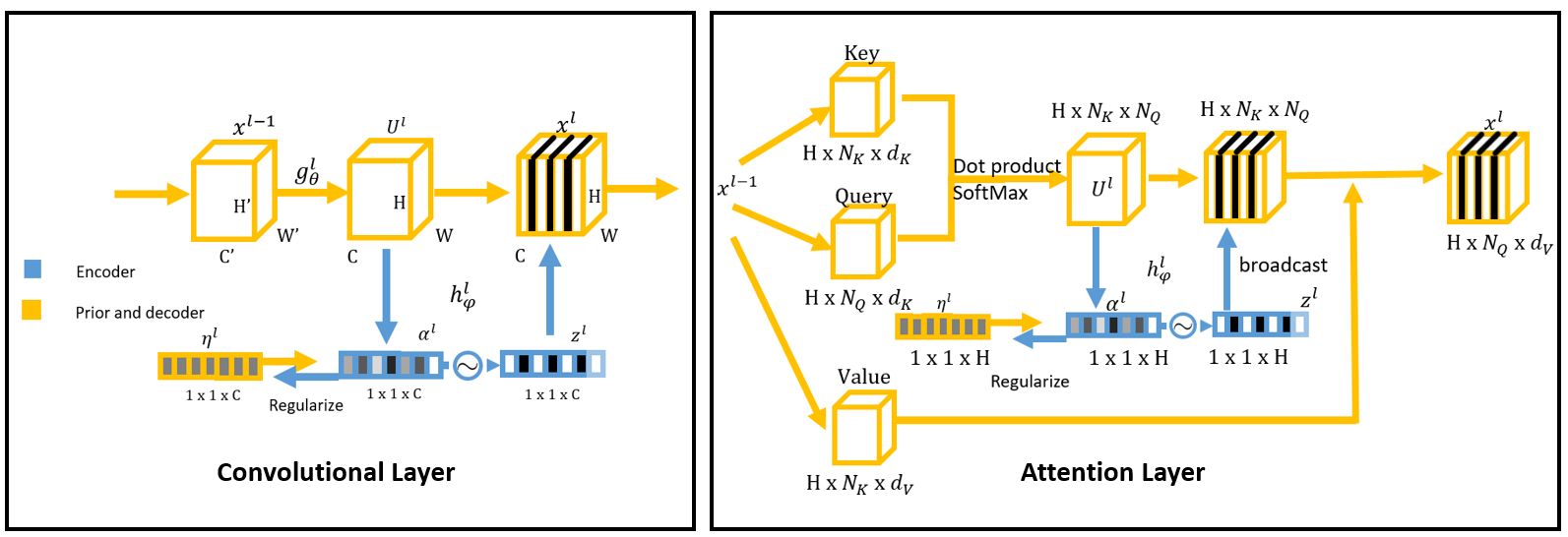} \vspace{-4mm}
\caption{\small \small {Left: Contextual dropout in convolution layers. Right: Contextual dropout in attention  layers.}} \label{fig:flow_conv_att}\vspace{-2mm}
\end{figure}

\textit{Contextual dropout module for convolutional layers:} 
Assume layer $l$ 
is a convolutional layer with $C_3^l$ as convolutional channels and $\mathbf{U}^l \in \mathbb{R}^{C_1^l\times C_2^l\times C_3^l}$.
Similar to Spatial Dropout \citep{tompson2015efficient}, we set $\alphav^l\in\mathbb{R}^{C_3^l}$ 
and broadcast its corresponding $\zv^l$ 
spatially as illustrated in Figure~\ref{fig:flow_conv_att}.
Such parameterization is similar to the squeeze-and-excitation unit for convolutional layers, which has been shown to be effective in image classification tasks \citep{hu2018squeeze}. {However, in squeeze-and-excitation, $\sigma(\alphav^l)$ is used as channel-wise soft attention weights instead of dropout probabilities, therefore it serves as a deterministic mapping in the model instead of a 
stochastic unit used in the inference network. 
}


\textit{Contextual dropout module for attention layers:} 
Dropout has been widely used in attention layers \citep{xu2015show,vaswani2017attention,yu2019deep}. For example, it can be applied to multi-head attention weights after the softmax operation (see illustrations in Figure~\ref{fig:flow_conv_att}). The weights are of dimension $[H, N_K, N_Q]$, where $H$ is the number of heads, $N_K$ the number of keys, and $N_Q$ the number of queries. In this case, we find that setting $\alphav^l\in\mathbb{R}^{H}$ 
gives good performance. Intuitively, this coincides with the choice of channel dimension for convolutional layers, as heads in attention 
could be analogized as channels in convolution.


\subsection{Variational inference for contextual dropout}
In contextual dropout, we choose $\mathcal{L}(\mathcal{D})=\sum_{(\xv,y)\in\mathcal{D}}\mathcal{L}(\xv,y)$ shown in \eqref{eq:ELBO_contextual} as the optimization objective.
Note 
in our design, 
the encoder $q_{\phivmat}$ reuses the decoder parameters $\thetav$ to define its own parameters. Therefore, we copy the values of $\thetav$ into $\phivmat$ and stop the gradient of $\thetav$ when optimizing~$q_{\phivmat}$.
This is theoretically sound \citep{ba2013adaptive}. Intuitively, the gradients to $\thetav$ from $p_{\thetav}$ are less noisy than that from $q_{\phiv}$
as the training of $p_{\thetav}(y\,|\,\xv, \zv)$ is supervised while that of $q_{\phiv}(\zv)$ is unsupervised. 
As what we have expected, allowing gradients from $q_{\phiv}$ to backpropagate to $\thetav$ 
is found to adversely 
affect the training of $p_{\thetav}$ 
in our experiments.
We use a simple prior $p_\etav$, making the prior distributions for dropout masks the same within each layer. 
The gradients 
with respect to $\etav$ and $\thetav$ can be expressed as
\ba{\nabla_{\etav}\mathcal{L}(\xv,y)=\E_{\zv\sim q_{\phivmat}(\cdot \given \xv)}[\nabla_{\etav} \log p _{\etav}(\zv)] 
, ~~~\nabla_{\thetav}\mathcal{L}(\xv,y)=\E_{\zv\sim q_{\phivmat}(\cdot \given \xv)}[\nabla_{\thetav} \log p _{\thetav}(y\given \xv,\zv)], }
which are both estimated via Monte Carlo integration, using a single $\zv\sim q_{\phivmat}(\zv \given \xv)$ for each $\xv$. 

%
%
%
%
Now, we consider the gradient of $\mathcal{L}$ with respect to $\varphiv$, the components of $\phivmat=\{\thetav, \varphiv\}$ 
not copied from the decoder. 
For Gaussian contextual dropout, we estimate the gradients via the reparameterization trick \citep{kingma2013auto}. For $\zv^l\sim N(\mathbf{1},\sigma(\alphav^{l})/(1-\sigma(\alphav^{l})))$, we rewrite it as $\zv^l = 1+\sqrt{\sigma(\alphav^{l})/(1-\sigma(\alphav^{l}))}\epsilonv^l$, where $\epsilonv^l\sim \mathcal N(\mathbf{0},\mathbf{I})$. 
Similarly, sampling a sequence of $\zv=\{\zv^l\}_{l=1}^L$ from $q_{\phiv}(\zv\given\xv)$ can be rewritten as $f_{\phiv}(\epsilonv,\xv)$, where $f_{\phiv}$ is a deterministic differentiable mapping and $\epsilonv$ are $iid$ standard Gaussian. The gradient $\nabla_{\varphiv}\mathcal{L}(\xv,y)$ can now be expressed as (see pseudo code of Algorithm~\ref{alg:reparameterization} in Appendix)
\ba{
\textstyle\nabla_{\varphiv}\mathcal{L}(\xv,y)=\E_{\epsilonv\sim \mathcal{N}(\mathbf{0},\mathbf{1})}[\nabla_{\varphiv} (\log p _{\thetav}(y\given \xv,f_{\phiv}(\epsilonv,\xv))-\frac{\log q_{\phiv}(f_{\phiv}(\epsilonv,\xv)\given\xv)}{\log p_\etav(f_{\phiv}(\epsilonv,\xv))})]. \label{eq:repara}}
For Bernoulli contextual dropout, backpropagating the gradient efficiently is not straightforward, as the Bernoulli distribution is not reparameterizable, restricting the use of the reparameterization trick. In this case, a commonly used gradient estimator is the REINFORCE estimator \citep{williams1992simple} (see details in Appendix~\ref{app:arm}). 
This estimator, however, is known to have high Monte Carlo estimation variance. 
%
To this end, we 
estimate $\nabla_{\varphiv} \mathcal L$ with the augment-REINFORCE-merge (ARM) estimator 
\citep{ARM}, which provides unbiased and low-variance gradients 
for the parameters of 
Bernoulli distributions. We defer the details of this estimator to Appendix~\ref{app:arm}. We note there exists an improved ARM estimator \citep{yin2020probabilistic,dong2020disarm}, applying which could further improve the performance. 

\subsection{Testing and complexity analysis}
{\bf Testing stage:} To obtain 
a point estimate, 
we follow the common practice in dropout \citep{srivastava2014dropout} 
to multiply the neurons by the expected values of random dropout masks, which means that we
predict $y$ with $p_{\thetav}(y\given \xv,\bar{\zv})$, where $\bar{\zv}=\E_{q_{\phiv}(\zv\given \xv)}[\zv]$ under the proposed contextual dropout. 
When uncertainty estimation is needed, we draw $K$ random dropout masks to approximate the posterior predictive distribution of $y$ given $\xv$ using
$\hat{p}(y\given \xv)=\frac{1}{K}\sum_{k=1}^K p_{\thetav}(y\given \xv,\zv^{(k)}),$ where $\zv^{(1)},\ldots,\zv^{(K)}\stackrel{iid}\sim q_{\phiv}(\zv\given \xv)$. 


{\bf Complexity analysis:} The added computation and memory of contextual dropout are insignificant due to the parameter sharing between the encoder and decoder. Extra memory and computational cost mainly comes from the part of $h_{\varphiv}^l$, where both the parameter size and number of operations are of order $O((C_d^l)^2/\gamma)$, where $\gamma$ is from $8$ to $16$. This is insignificant, compared to the memory and computational cost of the main network, which are of order larger than $O((C_d^l)^2)$. We verify the point by providing memory and runtime comparisons between contextual dropout and other dropouts on ResNet in Table~\ref{tab:resnet} (see more model size comparisons in Table \ref{table:parameter} in Appendix).

\subsection{
Related work}
{\it Data-dependent variational distribution:} 
\citet{deng2018latent} model attentions as latent-alignment variables and optimize a tighter lower bound (compared to hard attention) 
using a learned inference network. 
To balance exploration and exploitation for contextual bandits problems,
\citet{wang2019thompson} introduce local variable uncertainty under the Thompson sampling framework.  
However, their inference networks of are both independent of the decoder, 
which may considerably increase memory and computational cost for the considered applications. \citet{fan2020bayesian} propose Bayesian attention modules with efficient parameter sharing between the encoder and decoder networks. Its scope is limited to attention units as \citet{deng2018latent},
while we demonstrate the general applicability of contextual dropout to fully connected, convolutional, and attention layers in supervised learning  models. 
{\it Conditional computation} \citep{bengio2015conditional,bengio2013estimating,shazeer2017outrageously,teja2018hydranets} tries to increase model capacity without a proportional increase in computation, where an independent gating network decides turning which part of a network active and which inactive for each example. 
In contextual dropout, the encoder works much like a gating network choosing the distribution of sub-networks for each sample. But the potential gain in model capacity is even larger, $e.g.,$ there are potentially $\sim O((2^d)^L)$ combinations of nodes for $L$ fully-connected layers, where $d$ is the order of the number of nodes for one layer. {{\it Generalization of dropout}: DropConnect \citep{wan2013regularization} randomly drops the weights rather than the activations so as to generalize dropout. The dropout distributions for the weights, however, are still the same across different samples. Contextual dropout utilizes sample-dependent dropout probabilities, allowing different samples to have different dropout probabilities.}


\section{Experiments}



Our method can be straightforwardly deployed wherever regular dropout can be utilized. To test its general applicability and scalability, we apply the proposed method to three representative types of NN layers: fully connected, convolutional, and attention layers with applications on MNIST \citep{lecun2010mnist}, CIFAR \citep{krizhevsky2009learning}, ImageNet \citep{deng2009imagenet}, and VQA-v2 \citep{goyal2017making}. 
To investigate the model's robustness to noise, 
we also construct noisy versions of datasets by adding Gaussian noises to image inputs \citep{larochelle2007empirical}.  


For evaluation, we consider both the accuracy and uncertainty on predicting $y$ given $\xv$. Many metrics have been proposed to evaluate the quality of uncertainty estimation. On one hand, researchers are generating calibrated probability estimates 
to measure model confidence \citep{guo2017calibration,naeini2015obtaining,kuleshov2018accurate}. While expected calibration error 
and maximum calibration error 
have been proposed to quantitatively measure calibration, 
such metrics do not reflect how robust the probabilities are with noise injected into the network input, and cannot capture epistemic or model uncertainty \citep{gal2016dropout}. On the other hand, the entropy of the predictive distribution as well as the mutual information, between the predictive distribution and posterior over network weights, are used as metrics  to capture both epistemic and aleatoric uncertainty \citep{mukhoti2018evaluating}. However, 
it is often unclear 
how large the
entropy or mutual information is large enough to be classified as uncertain, so such metric only provides a relative uncertainty measure. 

{\bf Hypothesis testing based uncertainty estimation}: Unlike previous information theoretic metrics, we use a statistical test based method to estimate uncertainty, which works for both single-label and multi-label classification models. One advantage of using hypothesis testing over information theoretic metrics is that 
the $p$-value of the test 
can be more interpretable, 
making it easier to be deployed in practice to obtain a binary uncertainty decision. 
To quantify how confident our model is about this prediction, we evaluate whether the difference between the empirical distributions of the two most possible classes from multiple posterior samples 
is statistically significant. 
Please see Appendix~\ref{sec:two_sample} for a detailed explanation of the test procedure. 



{\bf Uncertainty evaluation via PAvPU:} With the $p$-value of the testing result and a given $p$-value threshold, we can determine whether the model is certain or uncertain about one prediction. To evaluate the uncertainty estimates, we uses Patch Accuracy vs Patch Uncertainty (PAvPU) \citep{mukhoti2018evaluating}, 
which is defined as 
 $\mathrm{PAvPU}={\left(n_{a c}+n_{i u}\right)}/{\left(n_{a c}+n_{a u}+n_{i c}+n_{i u}\right)}$, 
where $n_{ac}, n_{au}, n_{ic}, n_{iu}$ are the numbers of accurate and certain, accurate and uncertain, inaccurate and certain, inaccurate and uncertain samples, respectively. This PAvPU evaluation metric would be higher if the model tends to generate the accurate prediction with high certainty and inaccurate prediction with high uncertainty.

%
%
%
%

\subsection{Contextual dropout on fully connected layers}
\label{sec:exp_full}

We consider an MLP with two hidden layers of size $300$ and $100$, respectively, with ReLU activations. Dropout is applied to the input layer and the outputs of first two full-connected layers. We use MNIST as the benchmark. 
We compare contextual dropout with MC dropout \citep{gal2016dropout}, concrete dropout \citep{gal2017concrete}, Gaussian dropout \citep{srivastava2014dropout}, and Bayes by Backprop \citep{blundell2015weight}. 
Please see the detailed experimental setting in {Appendix}~\ref{sec:hyper_image}. 

\begin{table}[htp!]\vspace{-3mm}
\centering
\caption{\small Results on noisy MNIST with MLP.}
\label{table:mnist_noise1}
\centering
\begin{sc}\vspace{-2mm}
\resizebox{0.7\columnwidth}{!}{
\begin{tabular}{@{}llll@{}}\toprule
Methods & Accuracy & PAvPU\small{(0.05)} & log likelihood \\ \midrule
MC - Bernoulli 
& 86.36 & 85.63 & -1.72 \\ 
MC - Gaussian 
&  86.31&  85.64& -1.72\\
Concrete 
& 86.52& 86.77& -1.68 \\ 
Bayes By Backprop 
& 86.55 & 87.13& -2.30 \\ 
{Contextual gating} & 86.20 & -& -1.81 \\
{Contextual gating+Dropout} & 86.70 & 87.01 & -1.71 \\ 
\hline \hline \\ [-2.0ex]
Bernoulli Contextual 
& {\bf87.43}\small{$\pm0.39$} &  {\bf 87.81}\small{$\pm0.23$}& {\bf-1.41} \small{$\pm0.01$}\\
Gaussian Contextual 
& 87.35\small{$\pm0.33$} & 87.72\small{$\pm0.29$}&  -1.43\small{$\pm0.01$}\\ 
\bottomrule
\end{tabular}
}
\end{sc} 
\end{table}

\begin{figure}[tp!]\vspace{-2mm} \centering
\begin{subfigure}[t]{.25\textwidth}
 \centering
 \includegraphics[width=1\linewidth]{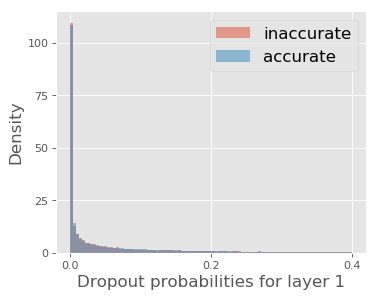}
 \vspace{1mm}
\end{subfigure}
\begin{subfigure}[t]{.252\textwidth}
 \centering
 \includegraphics[width=1\linewidth]{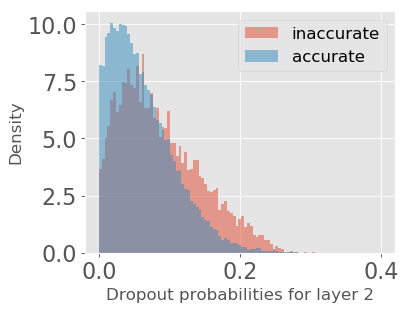}
\end{subfigure}
\begin{subfigure}[t]{.23\textwidth}
 \centering
 \includegraphics[width=1\linewidth]{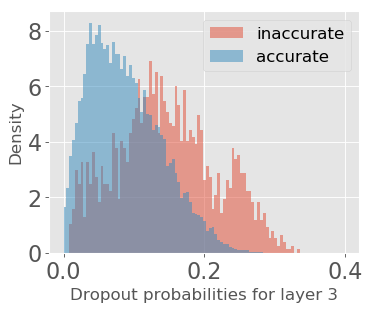}
 \vspace{1mm}
\end{subfigure}
\begin{subfigure}[t]{.24\textwidth}
 \centering
 \includegraphics[width=1\linewidth]{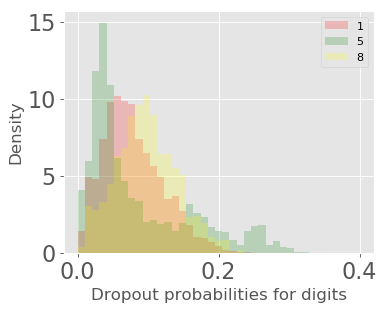}
\end{subfigure}
 \vspace{-5mm}
\caption{\small {Visualization of dropout probabilities of Bernoulli contextual dropout on the MNIST dataset}: the learned dropout probabilities seem to increase as we go to higher-level layers, as also observed in \citet{gal2017concrete}. With contextual dropout, different samples own different dropout probabilities. Inaccurate ones often have higher dropout probabilities corresponding to higher uncertainties. On the further right figure,  we compare the dropout distributions across $3$ representative digits. The dropout probabilities are overall higher for digit $8$ compared to digit $1$, meaning $1$ is easier to classify. The distribution for $5$ has longer tails than others showing there are more variations in the uncertainty for digit $5$.}
\label{fig:visual_dropout}\vspace{-3mm}
\end{figure}

{\bf Results and analysis:} 
In Table~\ref{table:mnist_noise1}, we show accuracy, PAvPU ($p$-value threshold equal to $0.05$)
and, 
test predictive loglikelihood with error bars ($5$ random runs) for models with different dropouts 
under the challenging noisy data\footnote{Results on original data is deferred to Table \ref{table:mnist_noise1_full} in Appendix 
.} (added Gaussian noise with mean $0$, variance $1$).
{Note that the uncertainty results for $p$-value threshold $0.05$ is 
in general 
consistent with the results for other $p$-value thresholds (see more in Table~\ref{table:mnist_noise1_full} in Appendix).} We observe that contextual dropout outperforms other methods in all metrics.
Moreover, compared to Bayes by Backprop, contextual dropout is more memory and computationally efficient. As shown in Table~\ref{table:parameter} in Appendix, contextual dropout only introduces $16\%$ additional parameters. However, Bayes by Backprop doubles the memory and increases the computations significantly as we need multiple draws of NN weights for uncertainty. Due to this reason, we do not include it for the following large model evaluations. 
{We note that using the output of the gating network to directly scale activations (contextual gating) underperforms contextual dropout, which shows that the sampling process is important for preventing overfitting and improving robustness to noise. Adding a regular dropout layer on the gating activations (contextual gating + dropout) improves a little, but still underperforms contextual dropout, demonstrating that how we use the gating activations matters. }
In Figure~\ref{fig:visual_dropout}, we observe that Bernoulli contextual dropout learns different dropout probabilities for different samples adapting the sample-level uncertainty which further verifies our motivation and supports the empirical improvements. {For sample-dependent dropout, the dropout probabilities would not vanish to zero even though the  prior for regularization is also learned, because the optimal dropout probabilities for each sample is not necessarily zero. Enabling different samples to have different network connections could greatly enhance the model’s capacity. The prior distribution also plays a different role here. Instead of preventing the dropout probabilities from going to zero, the prior tries to impose some similarities between the dropout probabilities of different samples.
}

\begin{figure}
    \centering
    \includegraphics[width=0.7\linewidth]{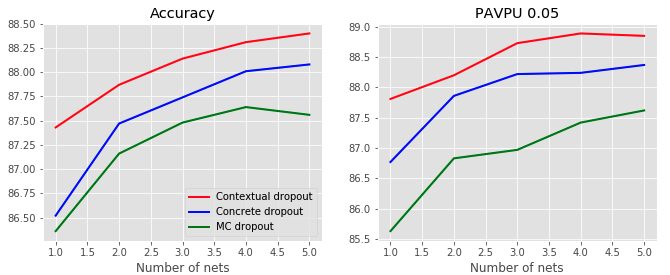}
     \caption{\small {The  performance of combining different dropouts with deep ensemble on noisy MNIST data}. }
 \label{fig:ensemble} \vspace{-7mm}
\end{figure}

\textbf{Combine contextual dropout with Deep Ensemble:} Deep ensemble proposed by \citet{lakshminarayanan2017simple} is a simple way to obtain uncertainty by ensembling models
trained
 independently from different random initializations. In Figure~\ref{fig:ensemble}, 
 we show the  performance 
of combining different  dropouts with deep ensemble on noisy MNIST data.
As the number of NNs increases, both accuracy and PAvPU increase for all dropouts. However, Bernoulli
  contextual  dropout outperforms 
other dropouts 
by a large margin in both metrics,  
showing 
contextual dropout is compatible  with deep ensemble and their combination can lead to significant 
improvements. 
 {\bf Out of distribution (OOD) evaluation: }we evaluate different dropouts in an OOD setting, where we train our model with clean data but test it on noisy data. Contextual dropout achieves accuracy of $78.08$, consistently higher than MC dropout ($75.22$) or concrete dropout ($74.93$). Meanwhile, the proposed method is also better at uncertainty estimation with PAvPU of $78.49$, higher than MC ($74.61$) or Concrete ($75.49$).



\subsection{Contextual dropout on convolutional layers}
We apply dropout to the convolutional layers in WRN 
\citep{zagoruyko2016wide}. In Figure \ref{fig:wrn_arch} in Appendix, we show the architecture of WRN, where dropout is applied to the first convolutional layer in each network block; in total, dropout is applied to $12$ convolutional layers.
%
We evaluate on CIFAR-10 and CIFAR-100 
. The detailed setting is provided in Appendix \ref{sec:hyper_image}. 

\begin{table}[htp!]\vspace{-3mm}
\centering
\caption{\small Results on CIFAR-100 with WRN.}\vspace{-2mm}
\label{table:CIFAR-10_100noise1}
\ra{0.5}
\resizebox{0.95\columnwidth}{!}{
\begin{tabular}{@{}llllclll@{}}\toprule
Dropout & \multicolumn{3}{c}{Original Data}& 
& \multicolumn{3}{c}{Noisy Data}\\
\cmidrule{2-4} \cmidrule{6-8}
& Accuracy & PAvPU \small{(0.05)}  & log likelihood && Accuracy & PAvPU \small{(0.05)}  & log likelihood \\ \midrule
Bernoulli & 79.03 & 61.54 & -4.49 && {52.01} & 54.25 & -4.55 \\
Gaussian &  76.63& 78.05 & -3.93 && {51.38} & 57.02 &  -4.23 \\
Concrete & {79.19} & 64.14 & -4.50 && 51.58& 56.61 & -4.56\\
\hline \hline \\ [-0.8ex]
\footnotesize{Bernoulli Contextual} & 80.85\small{$\pm0.05$} & 81.56\small{$\pm0.31$} & -3.56\small{$\pm0.02$} && 53.64\small{$\pm0.45$} & {\bf 58.63}\small{$\pm0.50$} & {\bf-3.73}\small{$\pm0.04$}\\
\footnotesize{Gaussian Contextual} & {\bf 80.93}\small{$\pm0.18$} & {\bf 81.69}\small{$\pm0.16$} & {\bf-3.43}\small{$\pm0.07$} &&  {\bf53.72}\small{$\pm0.34$} & 58.49\small{$\pm0.43$} & -3.81 \small{$\pm0.03$}\\
\bottomrule
\end{tabular}
}
\vspace{-2mm}
\end{table}

{\bf Results and analysis:} We show the results for 
CIFAR-100 in Table \ref{table:CIFAR-10_100noise1} (see CIFAR-10 results in Tables \ref{table:full_CIFAR-10_noise}-\ref{table:loglilikehood_CIFAR-10_100_noise} in Appendix). Accuracies, PAvPUs, and test predictive loglikelihoods 
are incorporated for 
both the original and noisy data. We consistently observe that contextual dropout outperforms other models in accuracy, uncertainty estimation, and loglikelihood.

\textbf{Uncertainty visualization:} We conducted extensive qualitative analyses for uncertainty evaluation. In Figures\! ~\ref{fig:CIFAR-10_concrete}
-
\ref{fig:CIFAR-10_rbnn}
in Appendix~\ref{sec:box_cifar10}, we visualize $15$ CIFAR images (with true label) and compare the corresponding probability outputs of different dropouts in boxplots. 
We observe (1) contextual dropout predicts the correct answer if it is certain, (2) contextual dropout is certain and predicts the correct answers on many images for which MC or concrete dropout is uncertain, (3) MC or concrete dropout is uncertain about some easy examples or certain on some wrong predictions (see details in Appendix~\ref{sec:box_cifar10}), (4) on an image that all three methods have high uncertainty, contextual dropout places a higher probability on the correct answer than the other two. These observations verify that contextual dropout provides better calibrated uncertainty. 

\begin{table}[htp!]\vspace{-3mm}
\centering
\caption{\small {Results on ImageNet with ResNet-18}. }
\label{tab:resnet}
\begin{sc}\vspace{-2mm}
\resizebox{0.75\linewidth}{!}{
\begin{tabular}{@{}llllll@{}}\toprule
Dropout & Top-1 Acc & PAvPU & Params & sec/step\\ \midrule
Without  & 69.75 & NA & 11.70M & 1.44\\
+Gaussian   & 69.46& 72.86 & 11.70M & 1.50\\
+Contextual  & {\bf70.03}\small{$\pm$0.07} & {\bf 74.68}\small{$\pm$0.08}&  11.88M & 1.64\\
+Contextual (scratch)  & {\bf70.29}\small{$\pm$0.09} & {\bf 76.47}\small{$\pm$0.12}&  11.88M & 1.64\\
\bottomrule
\end{tabular}}
\end{sc} 
\end{table}

{\bf Large-scale experiments with ImageNet:}
Contextual dropout is also applied to the convolutional layers in ResNet-18, where we plug contextual dropout into a pretrained model, and fine-tune the
pretrained model on ImageNet. In Table~\ref{tab:resnet}, we show it is even possible to finetune a pretrained model with contextual dropout module, and without much additional memory or run time cost, it achieves better performance than both the original model and the one with regular Gaussian dropout. Training model with contextual dropout from scratch can further improve the performance. See detailed experimental setting in Appendix \ref{sec:hyper_image}.

\subsection{Contextual dropout on attention layers}
\begin{figure*}[t]
 \centering
 \includegraphics[width=1\textwidth,height=4.60cm]{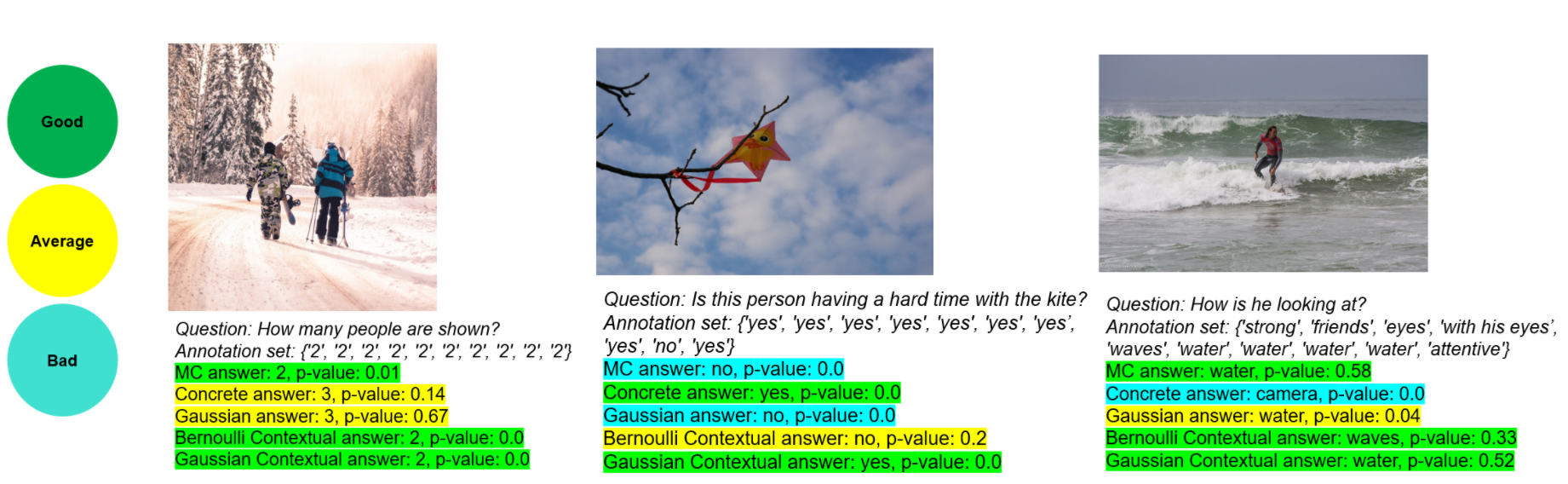}
 \vspace{-2.5mm}
 \caption{\small {VQA visualization}: for each question type, we present an image-question pair along with human annotations. We manually classify each prediction by different methods based on their answers and $p$-values.
For questions that have a clear answer, we define the good as certain \& accurate, the average as uncertain \& accurate or uncertain \& inaccurate, and the bad as certain \& inaccurate. Otherwise, we define the good as uncertain \& accurate, the average as certain \& accurate or uncertain \& inaccurate, and the bad as certain \& inaccurate. 
}
 \label{fig:vqa_vis_explanation} \vspace{-3mm}
\end{figure*}

We further apply contextual dropout to the attention layers of VQA models, whose goal is to provide an answer to a question relevant to the content of a given image. 
We conduct experiments on the commonly used benchmark, VQA-v2 \citep{goyal2017making}, containing human-annotated question-answer (QA) pairs.
There are three types of questions: Yes/No, Number, and Other. In Figure~\ref{fig:vqa_vis_explanation}, we show one example for each question type. There are $10$ answers provided by $10$ different human annotators for each question (see explanation of evaluation metrics in Appendix \ref{sec:exp_vqa}). {As shown in the examples, VQA is generally so challenging that there are often several different human annotations for a given image. 
Therefore, good uncertainty estimation becomes even more necessary.} 




{\bf Model and training specifications:}
We use MCAN \citep{yu2019deep}, a state-of-the-art {Transformer-like} model for VQA. Self-attention layers for question features and visual features, as well as the question-guided attention layers of visual features, are stacked one over another to build a deep model. Dropout is applied in every attention layer (after the softmax and before residual layer \citep{vaswani2017attention}) and fully-connected layer to prevent overfitting \citep{yu2019deep}, resulting in  $62$ dropout layers in total. {Experiments are conducted using the code of \citet{yu2019deep} as basis}. Detailed experiment setting is in Appendix~\ref{sec:exp_vqa}.

\begin{table}[htp!]
\caption{\small {Accuracy and PAvPU on visual question answering.}}\vspace{-2.5mm}
\label{tab:vqa_accuracy_uncer}
\centering
\ra{0.7}
\begin{sc}
\resizebox{0.9\columnwidth}{!}{
\begin{tabular}{@{}lllcll@{}}\toprule
Dropout & \multicolumn{2}{c}{Accuracy} & 
& \multicolumn{2}{c}{PAvPU}\\
\cmidrule{2-3} \cmidrule{5-6}
& Original Data & Noisy Data  && Original Data & Noisy Data \\ \midrule
{Bernoulli \citep{yu2019deep}} & {67.2} & - && - & - \\
MC - Bernoulli & {66.95} & 61.45 && 70.04 & 66.11 \\
MC - Gaussian & {66.96} & 62.75  && 70.77 & 67.42 \\
Concrete & 66.82 & 61.47  && 71.02 & 65.94  \\
\hline \hline \\ [-1.0ex]
Bernoulli Contextual & {\bf 67.42}\small{$\pm0.06$} & 63.73\small{$\pm0.08$} && {\bf71.65}\small{$\pm0.06$} & 68.57\small{$\pm0.11$} \\
Gaussian Contextual & {67.35}\small{$\pm0.03$} & {\bf63.82}\small{$\pm0.05$}  && 71.62\small{$\pm0.02$} & {\bf68.64}\small{$\pm0.04$}  \\
\bottomrule
\end{tabular}}
\end{sc} 
\vspace{-2.5mm}
\end{table}

{\bf Results and analysis:} We compare different dropouts on both the original VQA dataset and a noisy version, where Gaussian noise with standard deviation $5$ is added to the visual features. In Tables~\ref{tab:vqa_accuracy_uncer}, 
we show the overall accuracy and uncertainty estimation.
The results show that on the original data, contextual dropout achieves 
better accuracy and uncertainty estimation than the others. Moreover, on noisy data, where the prediction becomes more challenging and requires more model flexibility and robustness, contextual dropouts outperform their regular dropout counterparts by a large margin in terms of accuracy with consistent improvement across all three question types.

\textbf{Visualization: }In Figures~\ref{fig:vqa_vis_contextual}-\ref{fig:vqa_vis_random} in Appendix~\ref{sec:vqa_visualization}, we visualize some image-question pairs, along with the human annotations and compare the predictions and uncertainty estimations of different dropouts. 
We show three of them in Figure~\ref{fig:vqa_vis_explanation}.
As shown in the plots, overall contextual dropout is more conservative on its wrong predictions and more certain on its correct predictions than other methods 
(see more detailed explanations in Appendix~\ref{sec:vqa_visualization}).

\section{Conclusion}

We introduce contextual dropout as a simple and scalable data-dependent dropout module that 
achieves strong performance in both accuracy and uncertainty estimation on a variety of tasks including large scale applications.
With an efficient parameterization of the coviariate-dependent variational distribution, 
contextual dropout boosts the flexibility of Bayesian neural networks with only slightly increased memory and computational cost. 
We demonstrate the general applicability of contextual dropout on fully connected, convolutional, and attention layers, and also show that contextual dropout masks are compatible with both Bernoulli and Gaussian distribution. 
On both image classification and visual question answering tasks, contextual dropout consistently outperforms corresponding baselines. Notably, on ImageNet, we find it is possible to improve the performance of a pretrained model by adding the contextual dropout module during a finetuning stage.
Based on these results, we believe contextual dropout can serve as an efficient alternative to data-independent dropouts in the versatile tool box of dropout modules.

\section*{Acknowledgements}

The authors acknowledge the support of Grants IIS-1812699, IIS-1812641, ECCS-1952193, CCF-1553281, and CCF-1934904
from the U.S. National Science Foundation,  and the Texas Advanced Computing Center
for providing HPC resources that have contributed to
the research results reported within this paper. M. Zhou acknowledges the support of a gift fund from ByteDance Inc. 

\nocite{langley00}

{\small
\bibliography{reference}

\begin{thebibliography}{60}
\providecommand{\natexlab}[1]{#1}
\providecommand{\url}[1]{\texttt{#1}}
\expandafter\ifx\csname urlstyle\endcsname\relax
  \providecommand{\doi}[1]{doi: #1}\else
  \providecommand{\doi}{doi: \begingroup \urlstyle{rm}\Url}\fi

\bibitem[Abadi et~al.(2015)Abadi, Agarwal, Barham, Brevdo, Chen, Citro,
  Corrado, Davis, Dean, Devin, Ghemawat, Goodfellow, Harp, Irving, Isard, Jia,
  Jozefowicz, Kaiser, Kudlur, Levenberg, Man\'{e}, Monga, Moore, Murray, Olah,
  Schuster, Shlens, Steiner, Sutskever, Talwar, Tucker, Vanhoucke, Vasudevan,
  Vi\'{e}gas, Vinyals, Warden, Wattenberg, Wicke, Yu, and
  Zheng]{abadi2016tensorflow}
Mart\'{\i}n Abadi, Ashish Agarwal, Paul Barham, Eugene Brevdo, Zhifeng Chen,
  Craig Citro, Greg~S. Corrado, Andy Davis, Jeffrey Dean, Matthieu Devin,
  Sanjay Ghemawat, Ian Goodfellow, Andrew Harp, Geoffrey Irving, Michael Isard,
  Yangqing Jia, Rafal Jozefowicz, Lukasz Kaiser, Manjunath Kudlur, Josh
  Levenberg, Dan Man\'{e}, Rajat Monga, Sherry Moore, Derek Murray, Chris Olah,
  Mike Schuster, Jonathon Shlens, Benoit Steiner, Ilya Sutskever, Kunal Talwar,
  Paul Tucker, Vincent Vanhoucke, Vijay Vasudevan, Fernanda Vi\'{e}gas, Oriol
  Vinyals, Pete Warden, Martin Wattenberg, Martin Wicke, Yuan Yu, and Xiaoqiang
  Zheng.
\newblock {TensorFlow}: Large-scale machine learning on heterogeneous systems,
  2015.
\newblock URL \url{http://tensorflow.org/}.
\newblock Software available from tensorflow.org.

\bibitem[Ba \& Frey(2013)Ba and Frey]{ba2013adaptive}
Jimmy Ba and Brendan Frey.
\newblock Adaptive dropout for training deep neural networks.
\newblock In \emph{Advances in Neural Information Processing Systems}, pp.\
  3084--3092, 2013.

\bibitem[Bengio et~al.(2015)Bengio, Bacon, Pineau, and
  Precup]{bengio2015conditional}
Emmanuel Bengio, Pierre-Luc Bacon, Joelle Pineau, and Doina Precup.
\newblock Conditional computation in neural networks for faster models.
\newblock \emph{arXiv preprint arXiv:1511.06297}, 2015.

\bibitem[Bengio et~al.(2013)Bengio, L{\'e}onard, and
  Courville]{bengio2013estimating}
Yoshua Bengio, Nicholas L{\'e}onard, and Aaron Courville.
\newblock Estimating or propagating gradients through stochastic neurons for
  conditional computation.
\newblock \emph{arXiv preprint arXiv:1308.3432}, 2013.

\bibitem[Blei et~al.(2017)Blei, Kucukelbir, and McAuliffe]{blei2017variational}
David~M Blei, Alp Kucukelbir, and Jon~D McAuliffe.
\newblock Variational inference: A review for statisticians.
\newblock \emph{Journal of the American Statistical Association}, 112\penalty0
  (518):\penalty0 859--877, 2017.

\bibitem[Blundell et~al.(2015)Blundell, Cornebise, Kavukcuoglu, and
  Wierstra]{blundell2015weight}
Charles Blundell, Julien Cornebise, Koray Kavukcuoglu, and Daan Wierstra.
\newblock Weight uncertainty in neural networks.
\newblock \emph{arXiv preprint arXiv:1505.05424}, 2015.

\bibitem[Boluki et~al.(2020)Boluki, Ardywibowo, Dadaneh, Zhou, and
  Qian]{boluki2020learnable}
Shahin Boluki, Randy Ardywibowo, Siamak~Zamani Dadaneh, Mingyuan Zhou, and
  Xiaoning Qian.
\newblock Learnable {B}ernoulli dropout for {B}ayesian deep learning.
\newblock In \emph{Artificial Intelligence and Statistics}, 2020.

\bibitem[Deng et~al.(2009)Deng, Dong, Socher, Li, Li, and
  Fei-Fei]{deng2009imagenet}
Jia Deng, Wei Dong, Richard Socher, Li-Jia Li, Kai Li, and Li~Fei-Fei.
\newblock Imagenet: A large-scale hierarchical image database.
\newblock In \emph{2009 IEEE conference on computer vision and pattern
  recognition}, pp.\  248--255. Ieee, 2009.

\bibitem[Deng et~al.(2018)Deng, Kim, Chiu, Guo, and Rush]{deng2018latent}
Yuntian Deng, Yoon Kim, Justin Chiu, Demi Guo, and Alexander Rush.
\newblock Latent alignment and variational attention.
\newblock In \emph{Advances in Neural Information Processing Systems}, pp.\
  9712--9724, 2018.

\bibitem[Dong et~al.(2020)Dong, Mnih, and Tucker]{dong2020disarm}
Zhe Dong, Andriy Mnih, and George Tucker.
\newblock {DisARM}: An antithetic gradient estimator for binary latent
  variables.
\newblock In \emph{Advances in Neural Information Processing Systems 33}, 2020.

\bibitem[Fan et~al.(2020)Fan, Zhang, Chen, and Zhou]{fan2020bayesian}
Xinjie Fan, Shujian Zhang, Bo~Chen, and Mingyuan Zhou.
\newblock Bayesian attention modules.
\newblock \emph{Advances in Neural Information Processing Systems}, 33, 2020.

\bibitem[Gal \& Ghahramani(2016)Gal and Ghahramani]{gal2016dropout}
Yarin Gal and Zoubin Ghahramani.
\newblock Dropout as a {B}ayesian approximation: {R}epresenting model
  uncertainty in deep learning.
\newblock In \emph{international conference on machine learning}, pp.\
  1050--1059, 2016.

\bibitem[Gal et~al.(2017)Gal, Hron, and Kendall]{gal2017concrete}
Yarin Gal, Jiri Hron, and Alex Kendall.
\newblock Concrete dropout.
\newblock In \emph{Advances in Neural Information Processing Systems}, pp.\
  3581--3590, 2017.

\bibitem[Ghasemi \& Zahediasl(2012)Ghasemi and Zahediasl]{ghasemi2012normality}
Asghar Ghasemi and Saleh Zahediasl.
\newblock Normality tests for statistical analysis: A guide for
  non-statisticians.
\newblock \emph{International journal of endocrinology and metabolism},
  10\penalty0 (2):\penalty0 486, 2012.

\bibitem[Goyal et~al.(2017)Goyal, Khot, Summers-Stay, Batra, and
  Parikh]{goyal2017making}
Yash Goyal, Tejas Khot, Douglas Summers-Stay, Dhruv Batra, and Devi Parikh.
\newblock Making the v in vqa matter: Elevating the role of image understanding
  in visual question answering.
\newblock In \emph{Proceedings of the IEEE Conference on Computer Vision and
  Pattern Recognition}, pp.\  6904--6913, 2017.

\bibitem[Graves(2011)]{graves2011practical}
Alex Graves.
\newblock Practical variational inference for neural networks.
\newblock In \emph{Advances in neural information processing systems}, pp.\
  2348--2356, 2011.

\bibitem[Guo et~al.(2017)Guo, Pleiss, Sun, and Weinberger]{guo2017calibration}
Chuan Guo, Geoff Pleiss, Yu~Sun, and Kilian~Q Weinberger.
\newblock On calibration of modern neural networks.
\newblock In \emph{Proceedings of the 34th International Conference on Machine
  Learning-Volume 70}, pp.\  1321--1330. JMLR. org, 2017.

\bibitem[He et~al.(2015)He, Zhang, Ren, and Sun]{he2015delving}
Kaiming He, Xiangyu Zhang, Shaoqing Ren, and Jian Sun.
\newblock Delving deep into rectifiers: Surpassing human-level performance on
  imagenet classification.
\newblock In \emph{Proceedings of the IEEE international conference on computer
  vision}, pp.\  1026--1034, 2015.

\bibitem[Hinton et~al.(2012)Hinton, Srivastava, Krizhevsky, Sutskever, and
  Salakhutdinov]{hinton2012improving}
Geoffrey~E Hinton, Nitish Srivastava, Alex Krizhevsky, Ilya Sutskever, and
  Ruslan~R Salakhutdinov.
\newblock Improving neural networks by preventing co-adaptation of feature
  detectors.
\newblock \emph{arXiv preprint arXiv:1207.0580}, 2012.

\bibitem[Hochreiter \& Schmidhuber(1997)Hochreiter and
  Schmidhuber]{hochreiter1997long}
Sepp Hochreiter and J{\"u}rgen Schmidhuber.
\newblock Long short-term memory.
\newblock \emph{Neural computation}, 9\penalty0 (8):\penalty0 1735--1780, 1997.

\bibitem[Hoffman et~al.(2013)Hoffman, Blei, Wang, and
  Paisley]{hoffman2013stochastic}
Matthew~D Hoffman, David~M Blei, Chong Wang, and John Paisley.
\newblock Stochastic variational inference.
\newblock \emph{The Journal of Machine Learning Research}, 14\penalty0
  (1):\penalty0 1303--1347, 2013.

\bibitem[Hu et~al.(2018)Hu, Shen, and Sun]{hu2018squeeze}
Jie Hu, Li~Shen, and Gang Sun.
\newblock Squeeze-and-excitation networks.
\newblock In \emph{Proceedings of the IEEE conference on computer vision and
  pattern recognition}, pp.\  7132--7141, 2018.

\bibitem[Kingma \& Ba(2014)Kingma and Ba]{kingma2014adam}
Diederik~P Kingma and Jimmy Ba.
\newblock Adam: A method for stochastic optimization.
\newblock \emph{arXiv preprint arXiv:1412.6980}, 2014.

\bibitem[Kingma \& Welling(2013)Kingma and Welling]{kingma2013auto}
Diederik~P Kingma and Max Welling.
\newblock Auto-encoding variational {B}ayes.
\newblock \emph{arXiv preprint arXiv:1312.6114}, 2013.

\bibitem[Kingma et~al.(2015)Kingma, Salimans, and
  Welling]{kingma2015variational}
Durk~P Kingma, Tim Salimans, and Max Welling.
\newblock Variational dropout and the local reparameterization trick.
\newblock In \emph{Advances in Neural Information Processing Systems}, pp.\
  2575--2583, 2015.

\bibitem[Krizhevsky et~al.(2009)]{krizhevsky2009learning}
Alex Krizhevsky et~al.
\newblock Learning multiple layers of features from tiny images.
\newblock Technical report, Citeseer, 2009.

\bibitem[Kuleshov et~al.(2018)Kuleshov, Fenner, and
  Ermon]{kuleshov2018accurate}
Volodymyr Kuleshov, Nathan Fenner, and Stefano Ermon.
\newblock Accurate uncertainties for deep learning using calibrated regression.
\newblock \emph{arXiv preprint arXiv:1807.00263}, 2018.

\bibitem[Lakshminarayanan et~al.(2017)Lakshminarayanan, Pritzel, and
  Blundell]{lakshminarayanan2017simple}
Balaji Lakshminarayanan, Alexander Pritzel, and Charles Blundell.
\newblock Simple and scalable predictive uncertainty estimation using deep
  ensembles.
\newblock In \emph{Advances in Neural Information Processing Systems}, pp.\
  6402--6413, 2017.

\bibitem[Larochelle et~al.(2007)Larochelle, Erhan, Courville, Bergstra, and
  Bengio]{larochelle2007empirical}
Hugo Larochelle, Dumitru Erhan, Aaron Courville, James Bergstra, and Yoshua
  Bengio.
\newblock An empirical evaluation of deep architectures on problems with many
  factors of variation.
\newblock In \emph{Proceedings of the 24th international conference on Machine
  learning}, pp.\  473--480, 2007.

\bibitem[LeCun et~al.(2010)LeCun, Cortes, and Burges]{lecun2010mnist}
Yann LeCun, Corinna Cortes, and CJ~Burges.
\newblock {MNIST} handwritten digit database.
\newblock \emph{AT\&T Labs [Online]. Available: http://yann. lecun.
  com/exdb/mnist}, 2:\penalty0 18, 2010.

\bibitem[LeCun et~al.(2015)LeCun, Bengio, and Hinton]{lecun2015deep}
Yann LeCun, Yoshua Bengio, and Geoffrey Hinton.
\newblock Deep learning.
\newblock \emph{Nature}, 521\penalty0 (7553):\penalty0 436--444, 2015.

\bibitem[Li et~al.(2016)Li, Chen, Carlson, and Carin]{li2016preconditioned}
Chunyuan Li, Changyou Chen, David Carlson, and Lawrence Carin.
\newblock Preconditioned stochastic gradient {L}angevin dynamics for deep
  neural networks.
\newblock In \emph{Thirtieth AAAI Conference on Artificial Intelligence}, 2016.

\bibitem[Li \& Ji(2019)Li and Ji]{l0arm2019}
Yang Li and Shihao Ji.
\newblock {L0-ARM}: Network sparsification via stochastic binary optimization.
\newblock In \emph{The European Conference on Machine Learning (ECML)}, 2019.

\bibitem[Louizos \& Welling(2017)Louizos and
  Welling]{louizos2017multiplicative}
Christos Louizos and Max Welling.
\newblock Multiplicative normalizing flows for variational {B}ayesian neural
  networks.
\newblock In \emph{Proceedings of the 34th International Conference on Machine
  Learning-Volume 70}, pp.\  2218--2227. JMLR. org, 2017.

\bibitem[MacKay(1992)]{mackay1992practical}
David~JC MacKay.
\newblock A practical bayesian framework for backpropagation networks.
\newblock \emph{Neural computation}, 4\penalty0 (3):\penalty0 448--472, 1992.

\bibitem[Molchanov et~al.(2017)Molchanov, Ashukha, and
  Vetrov]{molchanov2017variational}
Dmitry Molchanov, Arsenii Ashukha, and Dmitry Vetrov.
\newblock Variational dropout sparsifies deep neural networks.
\newblock In \emph{Proceedings of the 34th International Conference on Machine
  Learning-Volume 70}, pp.\  2498--2507. JMLR. org, 2017.

\bibitem[Mukhoti \& Gal(2018)Mukhoti and Gal]{mukhoti2018evaluating}
Jishnu Mukhoti and Yarin Gal.
\newblock Evaluating bayesian deep learning methods for semantic segmentation.
\newblock \emph{arXiv preprint arXiv:1811.12709}, 2018.

\bibitem[Naeini et~al.(2015)Naeini, Cooper, and
  Hauskrecht]{naeini2015obtaining}
Mahdi~Pakdaman Naeini, Gregory Cooper, and Milos Hauskrecht.
\newblock Obtaining well calibrated probabilities using {B}ayesian binning.
\newblock In \emph{Twenty-Ninth AAAI Conference on Artificial Intelligence},
  2015.

\bibitem[Neal(2012)]{neal2012bayesian}
Radford~M Neal.
\newblock \emph{Bayesian learning for neural networks}, volume 118.
\newblock Springer Science \& Business Media, 2012.

\bibitem[Nesterov(1983)]{nesterov1983method}
Yurii~E Nesterov.
\newblock A method for solving the convex programming problem with convergence
  rate o (1/k\^{} 2).
\newblock In \emph{Dokl. akad. nauk Sssr}, volume 269, pp.\  543--547, 1983.

\bibitem[Pennington et~al.(2014)Pennington, Socher, and
  Manning]{pennington2014glove}
Jeffrey Pennington, Richard Socher, and Christopher Manning.
\newblock {Glove}: {G}lobal vectors for word representation.
\newblock In \emph{Proceedings of the 2014 conference on empirical methods in
  natural language processing (EMNLP)}, pp.\  1532--1543, 2014.

\bibitem[Ren et~al.(2015)Ren, He, Girshick, and Sun]{ren2015faster}
Shaoqing Ren, Kaiming He, Ross Girshick, and Jian Sun.
\newblock Faster {R-CNN}: Towards real-time object detection with region
  proposal networks.
\newblock In \emph{Advances in neural information processing systems}, pp.\
  91--99, 2015.

\bibitem[Rezende et~al.(2014)Rezende, Mohamed, and
  Wierstra]{rezende2014stochastic}
Danilo~Jimenez Rezende, Shakir Mohamed, and Daan Wierstra.
\newblock Stochastic backpropagation and approximate inference in deep
  generative models.
\newblock In \emph{ICML}, pp.\  1278--1286, 2014.

\bibitem[Shazeer et~al.(2017)Shazeer, Mirhoseini, Maziarz, Davis, Le, Hinton,
  and Dean]{shazeer2017outrageously}
Noam Shazeer, Azalia Mirhoseini, Krzysztof Maziarz, Andy Davis, Quoc Le,
  Geoffrey Hinton, and Jeff Dean.
\newblock Outrageously large neural networks: The sparsely-gated
  mixture-of-experts layer.
\newblock \emph{arXiv preprint arXiv:1701.06538}, 2017.

\bibitem[Shi et~al.(2018)Shi, Sun, and Zhu]{shi2018kernel}
Jiaxin Shi, Shengyang Sun, and Jun Zhu.
\newblock Kernel implicit variational inference.
\newblock In \emph{International Conference on Learning Representations}, 2018.
\newblock URL \url{https://openreview.net/forum?id=r1l4eQW0Z}.

\bibitem[Srivastava et~al.(2014)Srivastava, Hinton, Krizhevsky, Sutskever, and
  Salakhutdinov]{srivastava2014dropout}
Nitish Srivastava, Geoffrey Hinton, Alex Krizhevsky, Ilya Sutskever, and Ruslan
  Salakhutdinov.
\newblock Dropout: a simple way to prevent neural networks from overfitting.
\newblock \emph{The Journal of Machine Learning Research}, 15\penalty0
  (1):\penalty0 1929--1958, 2014.

\bibitem[Teja~Mullapudi et~al.(2018)Teja~Mullapudi, Mark, Shazeer, and
  Fatahalian]{teja2018hydranets}
Ravi Teja~Mullapudi, William~R Mark, Noam Shazeer, and Kayvon Fatahalian.
\newblock Hydranets: Specialized dynamic architectures for efficient inference.
\newblock In \emph{Proceedings of the IEEE Conference on Computer Vision and
  Pattern Recognition}, pp.\  8080--8089, 2018.

\bibitem[Teney et~al.(2018)Teney, Anderson, He, and Van
  Den~Hengel]{teney2018tips}
Damien Teney, Peter Anderson, Xiaodong He, and Anton Van Den~Hengel.
\newblock Tips and tricks for visual question answering: Learnings from the
  2017 challenge.
\newblock In \emph{Proceedings of the IEEE Conference on Computer Vision and
  Pattern Recognition}, pp.\  4223--4232, 2018.

\bibitem[Tompson et~al.(2015)Tompson, Goroshin, Jain, LeCun, and
  Bregler]{tompson2015efficient}
Jonathan Tompson, Ross Goroshin, Arjun Jain, Yann LeCun, and Christoph Bregler.
\newblock Efficient object localization using convolutional networks.
\newblock In \emph{Proceedings of the IEEE Conference on Computer Vision and
  Pattern Recognition}, pp.\  648--656, 2015.

\bibitem[Vaswani et~al.(2017)Vaswani, Shazeer, Parmar, Uszkoreit, Jones, Gomez,
  Kaiser, and Polosukhin]{vaswani2017attention}
Ashish Vaswani, Noam Shazeer, Niki Parmar, Jakob Uszkoreit, Llion Jones,
  Aidan~N Gomez, {\L}ukasz Kaiser, and Illia Polosukhin.
\newblock Attention is all you need.
\newblock In \emph{Advances in neural information processing systems}, pp.\
  5998--6008, 2017.

\bibitem[Wan et~al.(2013)Wan, Zeiler, Zhang, Le~Cun, and
  Fergus]{wan2013regularization}
Li~Wan, Matthew Zeiler, Sixin Zhang, Yann Le~Cun, and Rob Fergus.
\newblock Regularization of neural networks using dropconnect.
\newblock In \emph{International conference on machine learning}, pp.\
  1058--1066, 2013.

\bibitem[Wang \& Zhou(2019)Wang and Zhou]{wang2019thompson}
Zhendong Wang and Mingyuan Zhou.
\newblock Thompson sampling via local uncertainty.
\newblock \emph{arXiv preprint arXiv:1910.13673}, 2019.

\bibitem[Welling \& Teh(2011)Welling and Teh]{welling2011bayesian}
Max Welling and Yee~W Teh.
\newblock Bayesian learning via stochastic gradient langevin dynamics.
\newblock In \emph{Proceedings of the 28th international conference on machine
  learning (ICML-11)}, pp.\  681--688, 2011.

\bibitem[Williams(1992)]{williams1992simple}
Ronald~J Williams.
\newblock Simple statistical gradient-following algorithms for connectionist
  reinforcement learning.
\newblock In \emph{Reinforcement Learning}, pp.\  5--32. Springer, 1992.

\bibitem[Xu et~al.(2015{\natexlab{a}})Xu, Wang, Chen, and Li]{xu2015empirical}
Bing Xu, Naiyan Wang, Tianqi Chen, and Mu~Li.
\newblock Empirical evaluation of rectified activations in convolutional
  network.
\newblock \emph{arXiv preprint arXiv:1505.00853}, 2015{\natexlab{a}}.

\bibitem[Xu et~al.(2015{\natexlab{b}})Xu, Ba, Kiros, Cho, Courville,
  Salakhudinov, Zemel, and Bengio]{xu2015show}
Kelvin Xu, Jimmy Ba, Ryan Kiros, Kyunghyun Cho, Aaron Courville, Ruslan
  Salakhudinov, Rich Zemel, and Yoshua Bengio.
\newblock Show, attend and tell: Neural image caption generation with visual
  attention.
\newblock In \emph{International conference on machine learning}, pp.\
  2048--2057, 2015{\natexlab{b}}.

\bibitem[Yin \& Zhou(2018)Yin and Zhou]{ARM}
Mingzhang Yin and Mingyuan Zhou.
\newblock {ARM}: {A}ugment-{REINFORCE}-merge gradient for discrete latent
  variable models.
\newblock \emph{Preprint}, May 2018.

\bibitem[{Yin} et~al.(2020){Yin}, {Ho}, {Yan}, {Qian}, and
  {Zhou}]{yin2020probabilistic}
Mingzhang {Yin}, Nhat {Ho}, Bowei {Yan}, Xiaoning {Qian}, and Mingyuan {Zhou}.
\newblock {Probabilistic Best Subset Selection by Gradient-Based Optimization}.
\newblock \emph{arXiv e-prints}, 2020.

\bibitem[Yu et~al.(2019)Yu, Yu, Cui, Tao, and Tian]{yu2019deep}
Zhou Yu, Jun Yu, Yuhao Cui, Dacheng Tao, and Qi~Tian.
\newblock Deep modular co-attention networks for visual question answering.
\newblock In \emph{Proceedings of the IEEE Conference on Computer Vision and
  Pattern Recognition}, pp.\  6281--6290, 2019.

\bibitem[Zagoruyko \& Komodakis(2016)Zagoruyko and
  Komodakis]{zagoruyko2016wide}
Sergey Zagoruyko and Nikos Komodakis.
\newblock Wide residual networks.
\newblock \emph{arXiv preprint arXiv:1605.07146}, 2016.

\end{thebibliography}
\bibliographystyle{iclr2021_conference}
}
\newpage

\clearpage
\appendix


\begin{center}
  \Large{\bf 
  Appendix}  
\end{center}

\section{Details of ARM gradient estimator for Bernoulli contextual dropout}
\label{app:arm}
In this section, we will explain the implementation details of ARM for Bernoulli contextual dropout. To compute the gradients with respect to the parameters of the variational distribution,
a commonly used gradient estimator is the REINFORCE estimator \citep{williams1992simple} as 
\begin{equation}
 \begin{split}
 &\nabla_{\varphiv}\mathcal L(\xv,y) = \E_{\zv\sim q_{\phivmat}(\cdot \given \xv)}[r(\xv,\zv,y)\nabla_{\varphiv}\log q_{\phivmat}(\zv \given \xv)],~~\textstyle r(\xv,\zv,y): =\log \frac{ p_{\thetav}(y\given\xv,\zv) p_{\etav}(\zv)}{q_{{\phivmat}}(\zv\given\xv)}.
 \end{split}\notag
\end{equation}

This gradient estimator is, however, known to have high variance \citep{ARM}. To mitigate this issue, we use ARM to compute the gradient with Bernoulli random variable.

{\bf 
ARM gradient estimator: } In general, denoting $\sigma(\alphav) = 1 /(1+ e^{-\alphav}) $ as the sigmoid function, ARM expresses the gradient of $\mathcal{E}(\alphav)=
 \E_{\zv\sim{\prod_{k=1}^K \text{Ber}}(z_k;\sigma(\alphav_k))}[r(\zv)]$ 
 as
\ba{\nabla_{\alphav}\mathcal{E}(\alphav)= \E_{ \piv\sim\prod_{k=1}^K\text{Uniform}(\pi_k;0,1)}[g_{\text{ARM}}(\piv)],~~~g_{\text{ARM}}(\piv):=[ r(\zv_{\text{true}})-r(\zv_{\text{sudo}})](1/2-\piv),}
where $\zv_{\text{true}}:=\mathbf{1}_{[\piv<\sigma(\alphav)]}$ and $\zv_{\text{sudo}}:=\mathbf{1}_{[\piv>\sigma(-\alphav)]}$ are referred to as the true and pseudo actions, respectively, and $\mathbf{1}_{[\cdotv]}\in\{0,1\}^K$ is an indicator function.

{\bf Sequential ARM: } Note that the above equation is not directly applicable to our model due to the cross-layer dependence. However, the dropout masks within each layer are independent of each other conditioned on these of the previous layers, so we can break our expectation into a sequence and apply ARM sequentially. 
We rewrite $\mathcal{L} = \E_{\zv\sim q_{{\phivmat}}(\cdot \given \xv)} [r(\xv, \zv, y)]$. 
When computing $\nabla_{\varphiv}\mathcal L$, we can ignore the ${\varphiv}$ in $r$ 
as the expectation of 
$\nabla_{\varphiv} \log q_{\phivmat}(\zv\given \xv)$ 
is zero. 
Using the chain rule, we have $\nabla_{{\varphiv}} \mathcal L = \sum_{l=1}^L \nabla_{\alphav^l}\mathcal L \nabla_{{\varphiv}} \alphav^l$. With decomposition $\mathcal L = \E_{\zv^{1:l-1}\sim q_{{\phivmat}}(\cdot \given \xv)} \E_{\zv^{l}\sim \text{Ber}(\sigma (\alphav^l))} [r(\xv, \zv^{1:l}, y)]$,
where $r(\xv, \zv^{1:l}, y) : = \E_{\zv^{l+1:L}\sim q_{{\phivmat}}(\cdot \given \xv, \zv^{1:l})} [r(\xv, \zv, y)]$, we know 
\bas{
 &\small \nabla_{\alphav^l}\mathcal L = \E_{\zv^{1:l-1}\sim q_{{\phivmat}}(\cdot \given \xv)} \E_{\piv^l\sim\prod_{k}\text{Uniform}(\pi_k^l;0,1)} [g_{\text{ARM}}(\piv^l)],\\
 &\small g_{\text{ARM}}(\piv^l)=[ r(\xv, \zv^{1:l-1}, \zv^l_{\text{true}}, y)-r(\xv, \zv^{1:l-1},\zv^l_{\text{sudo}}, y)] (1/2-\piv^l),
}
where $\zv^l_{\text{true}}:=\mathbf{1}_{[\piv^l<\sigma(\alphav^l)]}$ and $\zv^l_{ \text{sudo}}:=\mathbf{1}_{[\piv^l>\sigma(-\alphav^l)]}$. We estimate the gradients via Monte Carlo integration. 
We provide the pseudo code in Algorithm~\ref{alg:arm}.




{\bf{Implementation details: }}
The computational complexity of sequential ARM is $O(L)$ times of that of the decoder computation.
Although  it is embarrassingly parallelizable, in practice, with limited computational resource available, it maybe be challenging to use sequential ARM when $L$ is fairly large. In such cases, the original non-sequential ARM can be viewed as an approximation 
to strike a good balance between efficiency and accuracy (see the pseudo code in Algorithm~\ref{alg:arm_independ} in Appendix). 
In our cases, for image classification models, $L$ is small enough ($3$ for MLP, $12$ for WRN) for us to use sequential ARM. For VQA, $L$ is as large as $62$ and hence we choose the non-sequential ARM.

To control the learning rate of the encoder, we use a scaled sigmoid function: $\sigma_t(\alphav^l) = \frac{1}{1+\exp(-t\alphav^l)}$, where a larger $t$ corresponding to a larger learning rate for the encoder. 
This function is also used in \citet{l0arm2019} to facilitate the transition of probability between $0$ and $1$ for the purpose of pruning NN weights.

\clearpage
\section{Algorithms}
Below, we present training algorithms for both Bernoulli and Gaussian contextual dropout.
\begin{algorithm}[htb!]
 \caption{ Bernoulli contextual dropout with sequential ARM}
 \label{alg:arm}
\begin{algorithmic}
 \STATE {\bfseries Input:} data $\mathcal{D}$, $r$, $\{g_{\thetav}^l\}_{l=1}^L$, $\{h_{\varphiv}^l\}_{l=1}^L$, step size $s$
 \STATE {\bfseries Output:} updated $\thetav$, $\varphiv$, $\etav$
 \REPEAT
 \STATE $G_{{\varphiv}} = 0$;
 \STATE Sample $\xv, y$ from data $\mathcal{D}$;
 \STATE $\xv^0 = \xv$
 \FOR{$l=1$ {\bfseries to} $L$}
 \STATE $U^{l} = g_{\thetav}^l(\xv^{l-1})$, $\alphav^l = h_{\varphiv}^l (U^{l})$
 \STATE Sample $\piv^l$ from Uniform(0,1);
 \STATE $\zv^l_{\text{true}}:=\mathbf{1}_{[\piv^l<\sigma_t(\alphav^l)]}$;
 \STATE $\zv^l_{\text{sudo}}:=\mathbf{1}_{[\piv^l>\sigma_t(-\alphav^l)]}$;
 \IF{$\zv^l_{\text{true}} =\zv^l_{\text{sudo}} $}
 \STATE $r_\text{sudo}^l =$None;
 \ELSE
 \STATE $\xv^{l}_\text{sudo} = U^{l} \odot\zv_{l, \text{sudo}}$
 \FOR{$k=l+1$ {\bfseries to} $L$}
 \STATE $U^{k}_\text{sudo} = g_{\thetav}^k(\xv^{k-1}_\text{sudo})$, $\alphav^k_\text{sudo} = h_{\varphiv}^k (U^{k}_\text{sudo} )$
 \STATE Sample $\piv^k_\text{sudo}$ from Uniform(0,1);
 \STATE $\zv^k_{\text{sudo}}:=\mathbf{1}_{[\piv^k_\text{sudo}<\sigma_t(\alphav^k_\text{sudo})]}$;
 \STATE $\xv^{k}_\text{sudo} = U^{k}_\text{sudo} \odot \zv_{k,\text{sudo}}$;
 \ENDFOR
 \STATE $r_\text{sudo}^l = r(\xv^L_\text{sudo}, y)$
 \ENDIF
 \STATE $\xv^{l} = U^{l} \odot\zv^l_{\text{true}}$
 \ENDFOR
 \STATE $r_\text{true} = r(\xv^L_\text{true}, y)$
 \FOR{$l=1$ {\bfseries to} $L$}
 \IF{$r_\text{sudo}^l$ is not None}
 \STATE $G_{{\varphiv}}= G_{{\varphiv}}+ t(r_\text{true} - r_\text{sudo}^l)(1/2-\piv^l) 
 \nabla_{{\varphiv}}\alphav^l$ ;
 \ENDIF
 \ENDFOR
 \STATE $\varphiv = \varphiv + s G_{{\varphiv}}$, with step-size $s$;
 \STATE $\thetav = \thetav + s \frac{\partial \log p_{\thetav}(y\given\xv,\zv_{1:L, \text{true}})}{\partial \thetav}$;
  \STATE $\etav = \etav + s \frac{\partial \log p_{\etav}(\zv_{1:L, \text{true}})}{\partial \etav}$;
 \UNTIL{convergence}
\end{algorithmic}
\end{algorithm}
\newpage
\begin{algorithm}[htb!]
 \caption{ Bernoulli contextual dropout with independent ARM}
 \label{alg:arm_independ}
\begin{algorithmic}
 \STATE {\bfseries Input:} data $\mathcal{D}$, $r$, $\{g_{\thetav}^l\}_{l=1}^L$, $\{h_{\varphiv}^l\}_{l=1}^L$, step size $s$
 \STATE {\bfseries Output:} updated $\thetav$, $\varphiv$, $\etav$
 \REPEAT
 \STATE $G_{{\varphiv}} = 0$;
 \STATE Sample $\xv, y$ from data $\mathcal{D}$;
 \STATE $\xv^0 = \xv$
 \FOR{$l=1$ {\bfseries to} $L$}
 \STATE $U^{l} = g_{\thetav}^l(\xv^{l-1})$, $\alphav^l = h_{\varphiv}^l (U^{l})$
 \STATE Sample $\piv^l$ from Uniform(0,1);
 \STATE $\zv^l_{\text{true}}:=\mathbf{1}_{[\piv^l<\sigma_t(\alphav^l)]}$;
 \STATE $\xv^{l} = U^{l} \odot\zv^l_{\text{true}}$
 \ENDFOR
 \STATE $r_\text{true} = r(\xv^L_\text{true}, y)$
 \STATE $\xv^0_{\text{sudo}} = \xv$
 \FOR{$l=1$ {\bfseries to} $L$}
 \STATE $U^{l}_{\text{sudo}} = g_{\thetav}^l(\xv^{l-1}_{\text{sudo}})$, $\alphav^l_{\text{sudo}} = h_{\varphiv}^l (U^{l}_{\text{sudo}})$
 \STATE $\zv^l_{\text{sudo}}:=\mathbf{1}_{[\piv^l_{\text{sudo}}>\sigma_t(-\alphav^l_{\text{sudo}})]}$;
 \STATE $\xv^{l}_{\text{sudo}} = U^{l}_{\text{sudo}} \odot\zv^l_{\text{sudo}}$
 \ENDFOR
 \STATE $r_\text{sudo} = r(\xv^L_\text{sudo}, y)$;
 \FOR{$l=1$ {\bfseries to} $L$}
 \STATE $G_{{\varphiv}}= G_{{\varphiv}}+ t(r_\text{true} - r_\text{sudo})(1/2-\piv^l) \nabla_{{\varphiv}}\alphav^l$ ; 
 \ENDFOR
 \STATE $\varphiv = \varphiv + s G_{{\varphiv}}$, with step-size $s$;
 \STATE $\thetav = \thetav + s \frac{\partial \log p_{\thetav}(y\given\xv,\zv_{1:L, \text{true}})}{\partial \thetav}$;
   \STATE $\etav = \etav + s \frac{\partial \log p_{\etav}(\zv_{1:L, \text{true}})}{\partial \etav}$;
 \UNTIL{convergence}
\end{algorithmic}
\end{algorithm}
\onecolumn 

\begin{algorithm}[htb!]
 \caption{  Gaussian contextual dropout with reparamaterization trick}
 \label{alg:reparameterization}
\begin{algorithmic}
 \STATE {\bfseries Input:} data $\mathcal{D}$, $r$, $\{g_{\thetav}^l\}_{l=1}^L$, $\{h_{\varphiv}^l\}_{l=1}^L$, step size $s$
 \STATE {\bfseries Output:} updated $\thetav$, $\varphiv$, $\etav$
 \REPEAT
 \STATE Sample $\xv, y$ from data $\mathcal{D}$;
 \STATE $\xv^0 = \xv$
 \FOR{$l=1$ {\bfseries to} $L$}
 \STATE $U^{l} = g_{\thetav}^l(\xv^{l-1})$, $\alphav^l = h_{\varphiv}^l (U^{l})$
 \STATE Sample $\epsilonv^l$ from $\mathcal{N}(0,1)$;
 \STATE $\tauv^l = \sqrt{\frac{1-\sigma_t(\alphav^l)}{\sigma_t(\alphav^l)}}$;
 \STATE $\zv^l:=\mathbf{1} + \tauv^l \odot\epsilonv^l$;
 \STATE $\xv^{l} = U^{l} \odot\zv^l$
 \ENDFOR

 \STATE $\varphiv = \varphiv + s \nabla_{\varphiv} (\log p _{\thetav}(y\given \xv,\zv_{1:L})-\frac{\log q_{\phiv}(\zv_{1:L}|\xv)}{\log p_\etav(\zv_{1:L})})$, with step-size $s$;
 \STATE $\thetav = \thetav + s \frac{\partial \log p_{\thetav}(y\given\xv,\zv_{1:L})}{\partial \thetav}$;
   \STATE $\etav = \etav + s \frac{\partial \log p_{\etav}(\zv_{1:L})}{\partial \etav}$;
 \UNTIL{convergence}
\end{algorithmic}
\end{algorithm}

\section{Details of Experiments}

All experiments are conducted using a single Nvidia Tesla V100 GPU.
\begin{table}[htp!]\centering
\begin{sc}
\caption{  Model size comparison among different methods.}
\label{table:parameter}\resizebox{0.5\columnwidth}{!}{
\begin{tabular}{@{}l|lllll@{}}\toprule
Method & MLP & WRN & MCAN & ResNet-18 \\ \midrule
MC or Concrete  & 267K & 36.5M & 58M & 11.6M \\ 
Contextual & 311K & 36.6M& 61M & 11.8M \\
Bayes By Backprop & 534K & -& - & - \\
\bottomrule
\end{tabular}} 
\end{sc}
\end{table}

{{\bf Choice of hyper-parameters in Contextual Dropout:} Contextual dropout introduces two additional hyperparameters compared to regular dropout. One is the channel factor $\gamma$ for the encoder network. In our experiments, the results are not sensitive to the choice of the value of the channel factor $\gamma$. Any number from 8 to 16 
would
give similar results, which is also observed in \citep{hu2018squeeze}. The other is the sigmoid scaling factor $t$ that controls the learning rate of the encoder. We find that the performance is not that sensitive to its value and it is often beneficial to make it smaller than the learning rate of the decoder. In all experiments considered in the paper, which cover various noise levels and model sizes, we have simply fixed it at $t=0.01$.}

\subsection{Image Classification}
\label{sec:hyper_image}
{\bf MLP: }We consider an MLP with two hidden layers of size $300$ and $100$, respectively, and use ReLU activations. Dropout is applied to all three full-connected layers. We use MNIST as the benchmark. All models are trained for $200$ epochs with batch size $128$ and the Adam optimizer \citep{kingma2014adam} ($\beta_1 = 0.9$, $\beta_2 = 0.999$). The learning rate is $0.001$. We compare contextual dropout with MC dropout \citep{gal2016dropout} and concrete dropout \citep{gal2017concrete}. For MC dropout, we use the hand-tuned dropout rate at $0.2$. For concrete dropout, we initialize the dropout rate at $0.2$ for Bernoulli dropout and the standard deviation parameter at 0.5 for Gaussian dropout. and set the Concrete temperature at $0.1$ \citep{gal2017concrete}. We initialize the weights in contextual dropout with \textit{He-initialization} preserving the magnitude of the variance of the weights in the forward pass \citep{he2015delving}. We initialize the biases in the way that the dropout rate is $0.2$ when the weights for contextual dropout are zeros. We also initialize our prior dropout rate at $0.2$. For 
hyperparameter tuning, 
we 
hold out 
$10,000$ samples randomly selected from the training set for validation. 
We use the chosen hyperparameters to train on the full training set ($60,000$ samples) and evaluate on the testing set ($10,000$ samples).
We use Leaky ReLU \citep{xu2015empirical} with $0.1$ as the non-linear operator in contextual dropout. The reduction ratio $\gamma$ is set as $10$, and sigmoid scaling factor $t$ as $0.01$. For Bayes by Backprop, we use $-\log \sigma_1=0, -\log \sigma_2=6, \pi=0.2$ (following the notation in the original paper). For evaluation, we set $M=20$.

\begin{figure}[!htp]
 \centering
 \includegraphics[width=0.4\textwidth,height=2.1cm]{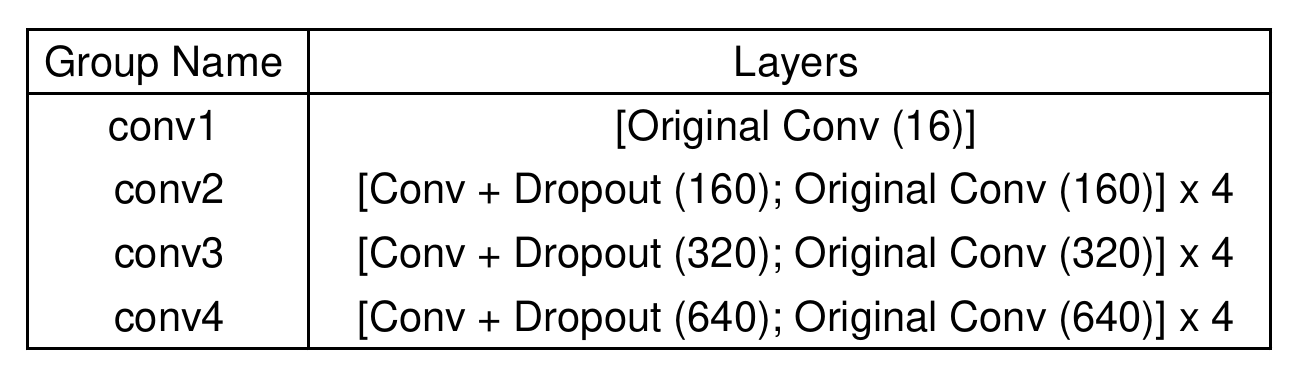}
 \caption{Architecture of the Wide Residual Network.}
 \label{fig:wrn_arch} 
\end{figure}

{\bf WRN: }We consider  WRN \citep{zagoruyko2016wide}, including $25$ convolutional layers. In Figure \ref{fig:wrn_arch}, we show the architecture of WRN, where dropout is applied to the first convolutional layer in each network block; in total, dropout is applied to $12$ convolutional layers.
We use CIFAR-10 and CIFAR-100 \citep{krizhevsky2009learning} as benchmarks. 
All experiments are trained for 200 epochs with the Nesterov Momentum optimizer \citep{nesterov1983method}, whose base learning rate is set as $0.1$, with decay factor $1/5$ at epochs 60 and 120. All other hyperparameters are the same as MLP except for Gaussian dropout, where we use standard deviation equal to 0.8 for the CIFAR100 with no noise and 1 for all other cases. 

{\bf ResNet: }We used ResNet-18 as the baseline model. We use momentum SGD, with learning rate $0.1$, and momentum weight $0.9$. Weight decay is utilized with weight $1e^{-4}$. For models trained from scratch, we train the models with $90$ epochs. For finetuning models, we start with pretrained baseline ResNet models and finetune for $1$ epoch. 

\subsection{VQA}
\label{sec:exp_vqa}
{\bf Dataset}: The dataset is split into the training (80k images and 444k QA pairs), validation (40k images and 214k QA pairs), and testing (80k images and 448k QA pairs) sets. We perform evaluation on the validation set as the true labels for the test set are not publicly available 
\citep{deng2018latent}. 

{\bf Evaluation metric:} the evaluation for VQA is different from image classification. The accuracy for a single answer could be a number between $0$ and $1$ \citep{goyal2017making}:
$\text{Acc}(ans) = \min \{{(\# \text{human that said } ans)}/{3},1\}.$
We generalize the uncertainty evaluation accordingly:
\bas{
\resizebox{0.45\hsize}{!}{$n_{ac} = \sum_{i} \text{Acc}_i \text{Cer}_i,~n_{iu} = \sum_{i} (1-\text{Acc}_i)(1- \text{Cer}_i)$\,}\notag, \resizebox{0.45\hsize}{!}{$n_{au} = \sum_{i} \text{Acc}_i (1-\text{Cer}_i),~n_{ic} = \sum_{i} (1-\text{Acc}_i)(\text{Cer}_i)$\,}
}
where  for the $i$th prediction $\text{Acc}_i$ is the accuracy and $\text{Cer}_i \in\{0,1\}$ is the  certainty indicator. 

{\bf Experimental setting:} We follow the setting by  \citet{yu2019deep}, where bottom-up features extracted from images by Faster R-CNN \citep{ren2015faster} are used as visual features, pretrained word-embeddings \citep{pennington2014glove} and LSTM \citep{hochreiter1997long} are used to extract question features.  We adopt the encoder-decoder structure in MCAN with six co-attention layers. We use the same model hyperparameters and training settings in \citet{yu2019deep} as follows:
the dimensionality of input image features, input question features, and fused multi-modal features are set to be $2048$, $512$, and $1024$, respectively. The latent dimensionality in the multi-head attention is $512$, the number of heads is set to $8$, and the latent dimensionality for each head is $64$. The size of the answer vocabulary is set to $N = 3129$ using the strategy in \citet{teney2018tips}. To train the MCAN model, we use the Adam optimizer \citep{kingma2014adam} with $\beta_1 = 0.9$ and $\beta_2 = 0.98$. The base learning rate is set to $\min(2.5te^{-5}, 1e^{-4})$, where $t$ is the current epoch number starting from $1$. After $10$ epochs, the learning rate is decayed by $1/5$ every $2$ epochs. All the models are trained up to $13$ epochs with the same batch size~of~$64$. 

We only conduct training on the training set (no data augmentation with visual genome dataset), and evaluation on the validation set. For MC dropout, we use the dropout rate of $0.1$ for Bernoulli dropout as in \citet{yu2019deep} and the standard deviation parameter of $1/3$ for Gaussian dropout. For concrete dropout, we initialize the dropout rate at $0.1$ and set the Concrete temperature at $0.1$ \citep{gal2017concrete}. For hyperparameter tuning, we randomly hold out $20\%$ of the training set for validation. After tuning, we train on the whole training set and evaluate on the validation set. We initialize the weights with \textit{He-initialization} preserving the magnitude of the variance of the weights in the forward pass \citep{he2015delving}. We initialize the biases in the way that the dropout rate is $0.1$ when the weights for contextual dropout are zeros. We also initialize our prior dropout rate at $0.1$. We use ReLU as the non-linear operator in contextual dropout. We use 
$\gamma=8$ for layers with $C_d^l > 8$, otherwise $\gamma = 1$. We set $\alpha \in \mathbb{R}^{d_V}$ for residual layers.

\section{Statistical test for uncertainty estimation}
\label{sec:two_sample}
\begin{figure}[!htb]
\begin{subfigure}[t]{.25\textwidth}
 \centering
 \includegraphics[width=1\linewidth]{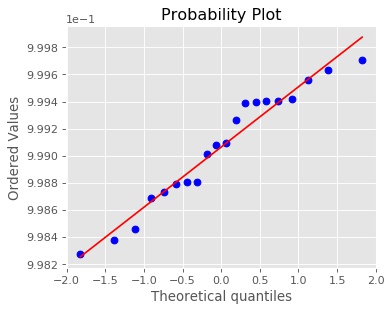}
 \vspace{1mm}
\end{subfigure}
\begin{subfigure}[t]{.25\textwidth}
 \centering
 \includegraphics[width=1\linewidth]{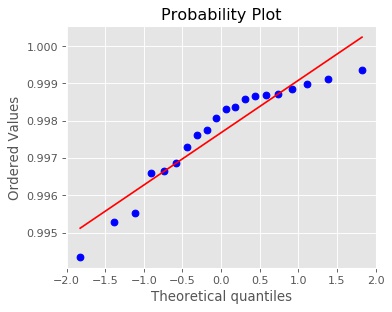}
 \vspace{-3mm}
\end{subfigure}
\begin{subfigure}[t]{.25\textwidth}
 \centering
 \includegraphics[width=1\linewidth]{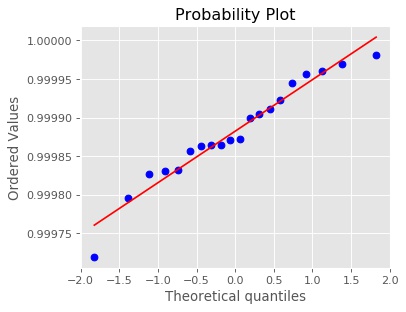}
 \vspace{-3mm}
\end{subfigure}
\begin{subfigure}[t]{.25\textwidth}
 \centering
 \includegraphics[width=1\linewidth]{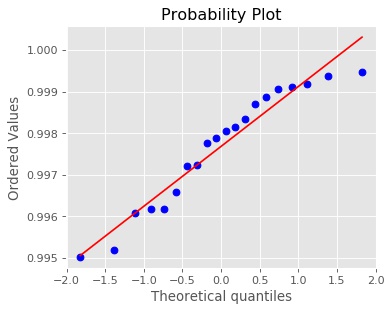}
 \vspace{-3mm}
\end{subfigure}\vspace{-3mm}\\
\begin{subfigure}[t]{.25\textwidth}
 \centering
 \includegraphics[width=1\linewidth]{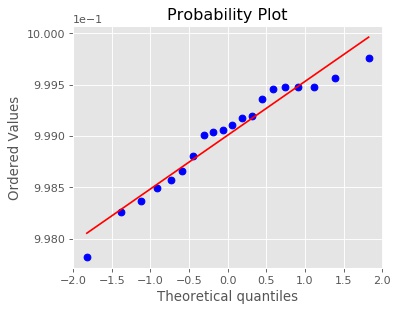}
 \vspace{1mm}
\end{subfigure}
\begin{subfigure}[t]{.25\textwidth}
 \centering
 \includegraphics[width=1\linewidth]{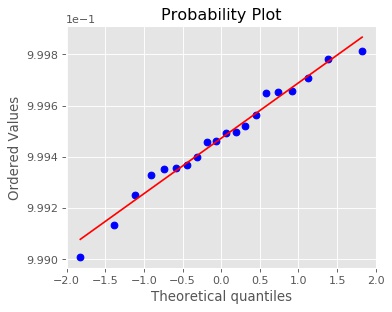}
 \vspace{-3mm}
\end{subfigure}
\begin{subfigure}[t]{.25\textwidth}
 \centering
 \includegraphics[width=1\linewidth]{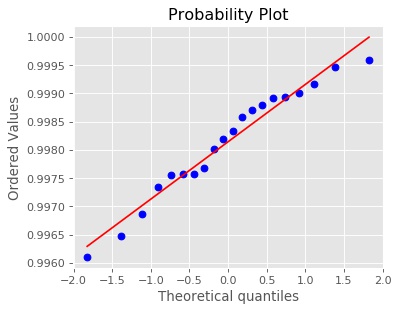}
 \vspace{-3mm}
\end{subfigure}
\begin{subfigure}[t]{.25\textwidth}
 \centering
 \includegraphics[width=1\linewidth]{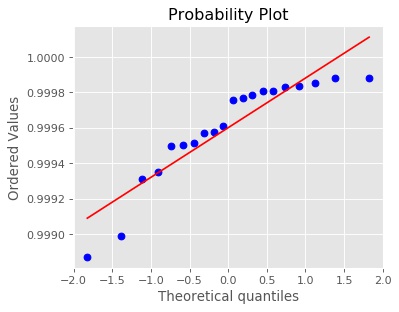}
 \vspace{-3mm}
\end{subfigure}%
\caption{  QQ Plot for differences between highest probabilities and second highest probabilities on WRN model with contextual dropout trained on the orignal CIFAR-10 dataset.}
\label{fig:qq_imageclass}\vspace{-3mm}
\end{figure}

\begin{figure}[!htb]
\begin{subfigure}[t]{.25\textwidth}
 \centering
 \includegraphics[width=1\linewidth]{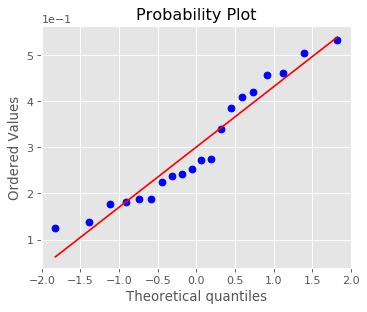}
 \vspace{1mm}
\end{subfigure}
\begin{subfigure}[t]{.25\textwidth}
 \centering
 \includegraphics[width=1\linewidth]{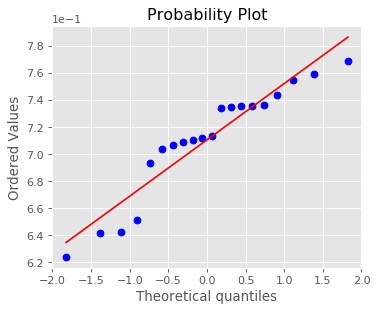}
 \vspace{-3mm}
\end{subfigure}
\begin{subfigure}[t]{.25\textwidth}
 \centering
 \includegraphics[width=1\linewidth]{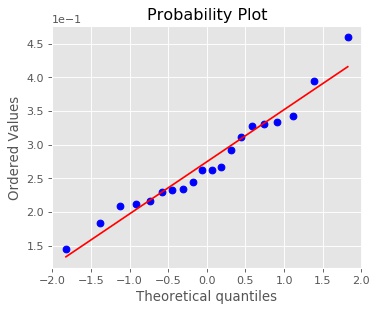}
 \vspace{-3mm}
\end{subfigure}
\begin{subfigure}[t]{.25\textwidth}
 \centering
 \includegraphics[width=1\linewidth]{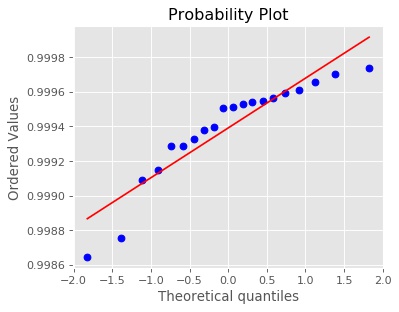}
 \vspace{-3mm}
\end{subfigure}\vspace{-3mm}\\
\begin{subfigure}[t]{.25\textwidth}
 \centering
 \includegraphics[width=1\linewidth]{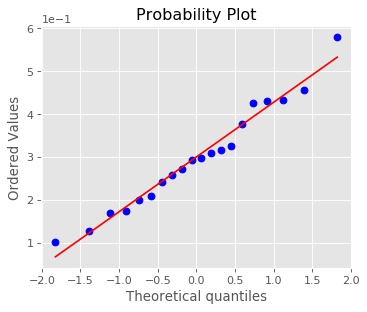}
 \vspace{1mm}
\end{subfigure}
\begin{subfigure}[t]{.25\textwidth}
 \centering
 \includegraphics[width=1\linewidth]{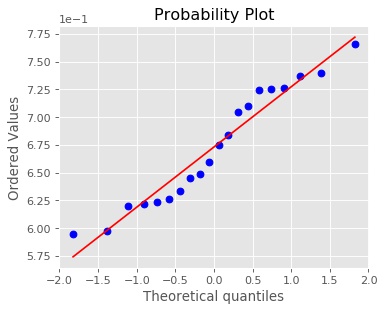}
 \vspace{-3mm}
\end{subfigure}
\begin{subfigure}[t]{.25\textwidth}
 \centering
 \includegraphics[width=1\linewidth]{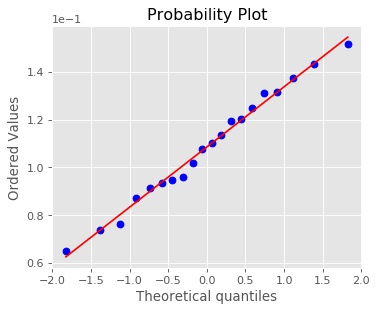}
 \vspace{-3mm}
\end{subfigure}
\begin{subfigure}[t]{.25\textwidth}
 \centering
 \includegraphics[width=1\linewidth]{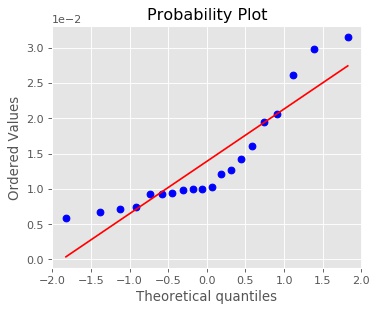}
 \vspace{-3mm}
\end{subfigure}%
\caption{  QQ Plot for output probabilities of VQA models: top row corresponds to the probability distributions of the class with the highest probability, and the bottom row corresponds to the probability distributions of the class with the second highest probability.}
\label{fig:qq_vqa}\vspace{-3mm}
\end{figure}

Consider $M$ posterior samples of predictive probabilities $\{\pv_m\}_{m=1}^M$, where $\pv_m$ is a vector with the same dimension as the 
number of classes. For single-label classification models, $\pv_m$ is produced by a softmax layer and sums to one, while for multi-label classification models, $\pv_m$ is produced by a sigmoid layer and each element is between $0$~and~$1$. The former output is used in most image classification models, while the latter is often used in VQA where multiple answers could be true for a single input. In both cases, 
to quantify how confident our model is about this prediction, we evaluate whether the difference between the probabilities of the first and second highest classes 
is statistically significant with a statistical test. We conduct the normality test on the output probabilities for both image classification  and VQA models, and find most of the output probabilities are approximately normal (we randomly pick some Q-Q plots \citep{ghasemi2012normality} and show them in Figures \ref{fig:qq_imageclass} and \ref{fig:qq_vqa}). This motivates us to use two-sample t-test\footnote{Note that we also tried a nonparametric test, Wilcoxon rank-sum test, and obtain similar results.}. In the following, we briefly summarize the two-sample $t$-test we use. 

Two sample hypothesis testing is an inferential statistical test that determines whether there is a statistically significant difference between the means in two groups. The null hypothesis for the $t$-test is that the population means from the two groups are equal: $\mu_1 = \mu_2$, and the alternative hypothesis is $\mu_1 \neq \mu_2$.  Depending on whether each sample in one group can be paired with another sample in the other group, we have either paired $t$-test or independent 
$t$-test. In our experiments, we utilize both types of two sample $t$-test. For a single-label model, the probabilities are dependent between two classes due to the softmax layer, therefore, we use the \textit{paired} two-sample $t$-test; for a multi-label model, the probabilities are independent given the logits of the output layer, so we use the \textit{independent} two-sample $t$-test.

For paired two-sample $t$-test, we calculate the difference between the paired observations 
calculate the $t$-statistic as below:
$$
T=\frac{\bar{Y}}{s / \sqrt{N}},
$$
where $\bar{Y}$ is the mean difference between the paired observations, $s$ is the standard deviation of the differences, and $N$ is the number of observations. Under the null hypothesis, this statistic follows a $t$-distribution with $N-1$ degrees of freedom if the difference is normally distributed. Then, we use this $t$-statistic and $t$-distribution to calculate the corresponding $p$-value.

For independent two-sample $t$-test, we calculate the  $t$-statistic as below:
$$
T=\frac{\bar{Y}_{1}-\bar{Y}_{2}}{\sqrt{s^{2} / N_{1}+s^{2} / N_{2}}}$$
$$s^2 = \frac{\sum (y_1-\bar{Y}_1) +\sum (y_2 - \bar{Y}_2)}{N_1+N_2-2}
$$
where $N_1$ and $N_2$ are the sample sizes, and $\bar{Y}_1$ and $\bar{Y}_2$ are the sample means. Under the null hypothesis, this statistic follows a $t$-distribution with $N_1 + N_2-2$ degrees of freedom if both $y_1$ and $y_2$ are normally distributed. We calculate the $p$-value accordingly.

To justify the assumption of the two-sample $t$-test, we run the normality test on the output probabilities for both image classification  and VQA models. We find most of the output probabilities are approximately normal. We randomly pick some Q-Q plots \citep{ghasemi2012normality} and show them in Figures \ref{fig:qq_imageclass} and \ref{fig:qq_vqa}. 

\section{Tables and Figures for p-value 0.01, 0.05 and 0.1}
\label{app:table}

\begin{table}[!htp]
\caption{  Complete results on MNIST with MLP}
\label{table:mnist_noise1_full}
\centering
\ra{0.7}
\begin{sc}
\resizebox{\textwidth}{!}{%
\begin{tabular}{@{}lllcll@{}}\toprule
& \multicolumn{2}{c}{Original Data} & \phantom{abc}& \multicolumn{2}{c}{Noisy Data}\\
\cmidrule{2-3} \cmidrule{5-6}
& Accuracy & PAvPU\small{(0.01 / 0.05 / 0.1)} && Accuracy & PAvPU\small{(0.01 / 0.05 / 0.1)} \\ \midrule
MC dropout - Bernoulli & 98.62 & 98.25 / 98.39 / 98.44 && 86.36 & 84.29/ 85.63 / 86.10 \\
MC dropout - Gaussian & 98.67 & 98.23 / 98.41/ 98.46 && 86.31 & 83.99 / 85.64 / 86.03 \\
Concrete dropout & 98.61 & 98.43/ 98.50 / 98.57 && 86.52 & 85.98 / 86.77/ 86.92\\
Bayes By Backprop 
& 98.44 &  98.26 / 98.42 / 98.56 && 86.55 & 86.89/ 87.13/ 87.26\\ 
\hline \hline \\ [-1.0ex]
Bernoulli Contextual Dropout & {\bf 99.08}\small{(0.04)} & {\bf 98.74}\small{(0.17)} / {\bf 98.92}\small{(0.08)} / {\bf 99.09}\small{(0.08)} && {\bf 87.43}\small{(0.39)} & {\bf 87.75}\small{(0.24)} / {\bf 87.81}\small{(0.23)} / {\bf 87.89}\small{(0.25)} \\
Gaussian Contextual Dropout & 98.92\small{(0.09)} & 98.71\small{(0.02)} / 98.90\small{(0.08)} / 99.03\small{(0.07)} && 87.35\small{(0.33)} & 87.64\small{(0.19)} / 87.72\small{(0.29)} / 87.78\small{(0.32)} \\
\bottomrule 
\end{tabular}
}
\end{sc}
\end{table}

\vspace{-2mm}
\begin{table}[htp!]
\vspace{-0.1in}
\caption{\small Loglikelihood on original MNIST with MLP.}
\label{table:mnist_original_log}\vspace{-0.007in}
\centering
\begin{sc}
\ra{0.7}
\resizebox{0.5\columnwidth}{!}{
\begin{tabular}{@{}ll@{}}\toprule
&  log likelihood \\ \midrule
MC - Bernoulli 
&  -1.4840 \small{$\pm0.0004$}\\ 
MC - Gaussian 
&  -1.4820\small{$\pm0.0003$}\\
Concrete 
&  -1.4822 \small{$\pm0.0012$}\\ 
Bayes By Backprop 
& -1.4806 \small{$\pm0.0007$}\\ 
\hline \hline \\ [-1.0ex]
Bernoulli Contextual 
& {\bf-1.4537} \small{$\pm0.0005$} \\
Gaussian Contextual 
& -1.4589\small{$\pm0.0005$} \\ 
\bottomrule
\end{tabular}
}
\end{sc}\vspace{-2mm}
\end{table}

\begin{table}[!htp]\centering
\caption{  Complete results on CIFAR-10 with WRN}
\label{table:full_CIFAR-10_noise}
\ra{0.7}
\begin{sc}
\resizebox{\textwidth}{!}{%
\begin{tabular}{@{}lllcll@{}}\toprule
& \multicolumn{2}{c}{Original Data} & \phantom{abc}& \multicolumn{2}{c}{Noisy Data}\\
\cmidrule{2-3} \cmidrule{5-6}
& Accuracy & PAvPU\small{(0.01 / 0.05 / 0.1)} && Accuracy & PAvPU\small{(0.01 / 0.05 / 0.1)} \\ \midrule
MC dropout - Bernoulli & 94.58& 78.73 / 82.34 / 84.21&& 79.51 & 72.89 / 74.43 / 75.04 \\
MC dropout - Gaussian & 93.81  & 92.59 / 93.24 / 93.85 && 79.33 & 80.43 / 81.24 / 82.31 \\
Concrete dropout & 94.60 & 73.51 / 78.41 / 81.01 && 79.34 & 72.72 / 73.89 / 74.72 \\
\hline \hline \\ [-1.0ex]
Bernoulli Contextual Dropout &  95.92\small{(0.10)} &  95.25\small{(0.23)} /  95.74\small{(0.12)} /  96.02\small{(0.16)} && 81.49\small{(0.19)} &  {\bf82.56}\small{(0.50)} /  {\bf83.28}\small{(0.31)} /  {\bf83.91}\small{(0.28)} \\
Gaussian Contextual Dropout & {\bf 96.04}\small{(0.1)} & {\bf 95.42}\small{(0.07)} / {\bf 95.85}\small{(0.07)} / {\bf 96.10}\small{(0.06)} && {\bf 81.64}\small{(0.31)} & 82.38 \small{(0.41)} / 82.80\small{(0.36)} / 83.43\small{(0.36)} \\
\bottomrule
\end{tabular}
}
\end{sc}
\end{table}

\begin{table}\centering
\ra{0.7}
\caption{ Complete log likelihood results on CIFAR-10 with WRN}
\label{table:loglilikehood_CIFAR-10_100_noise}
\begin{sc}
\resizebox{0.7\textwidth}{!}{%
\begin{tabular}{@{}lll@{}}\toprule
& \multicolumn{2}{c}{Cifar-10} \\
\cmidrule{2-3} 
& Original data & Noisy data \\ \midrule
MC dropout - Bernoulli & -1.91 & -1.93\\
MC dropout - Gaussian & -1.54 & -1.72   \\
Concrete dropout & -1.98  & -2.0 \\
\hline \hline \\ [-1.0ex]
Bernoulli Contextual Dropout & -1.24 & {\bf-1.47}\\
Gaussian Contextual Dropout & {\bf-1.19} &-1.51\\
\bottomrule
\end{tabular}
}
\end{sc}
\end{table}

\begin{table}\centering
\ra{0.7}
\caption{  Complete results on CIFAR-100 with WRN}
\label{table:full_CIFAR-100_noise}
\begin{sc}
\resizebox{\textwidth}{!}{%
\begin{tabular}{@{}lllcll@{}}\toprule
& \multicolumn{2}{c}{Original Data} & & \multicolumn{2}{c}{Noisy Data}\\
\cmidrule{2-3} \cmidrule{5-6}
& Accuracy & PAvPU\small{(0.01 / 0.05 / 0.1)} && Accuracy & PAvPU\small{(0.01 / 0.05 / 0.1)} \\ \midrule
MC dropout - Bernoulli & 79.03 & 56.90 / 61.54 / 64.14 &&  52.01 & 53.86 / 54.25 / 54.63 \\
MC dropout - Gaussian & 76.63 & 77.35 / 78.05 / 78.26 && 51.38& 56.83 / 57.02 / 57.31 \\
Concrete dropout & 79.19 & 59.45 / 64.14/ 66.63 && 51.58& 57.62 / 56.61/ 55.89\\
\hline \hline \\ [-1.0ex]
Bernoulli Contextual Dropout & 80.85\small{(0.05)} & 81.04\small{(0.28)} /  81.56\small{(0.31)} /  81.86\small{(0.21)} && 53.64\small{(0.45)} & {\bf 58.29}\small{(0.30)} / {\bf 58.63}\small{(0.50)} / {\bf 59.36}\small{(0.49)} \\
Gaussian Contextual Dropout & {\bf80.93} \small{(0.18)} & {\bf81.43}\small{(0.1)} / {\bf 81.69}\small{(0.16)} / {\bf 82.02}\small{(0.14)} && \bf{53.72}\small{(0.34)} & 58.01\small{(0.6)} / 58.49\small{(0.43)} / 58.95\small{(0.37)} \\
\bottomrule
\end{tabular}
}
\end{sc}
\end{table}

\section{Qualitative Analysis}
In this section, we include the Q-Q plots of the output probabilities as the normality test for the assumptions of two-sample $t$-test. In Figure~\ref{fig:qq_imageclass}, we test the normality of differences between highest probabilities and second highest probabilities on WRN model with contextual dropout trained on the orignal CIFAR-10 dataset. In Figure~\ref{fig:qq_vqa}, we test the normality of highest probabilities and second highest probabilities (separately) on VQA model with contextual dropout trained on the original VQA-v2 dataset. We use $20$ data points for the plots.

\subsection{Normality test of output probabilities}

\clearpage
\subsection{Boxplot for CIFAR-10}
\label{sec:box_cifar10}
\begin{figure*}[h]

\begin{subfigure}[t]{.25\textwidth}
 \centering
 \includegraphics[width=0.8\linewidth]{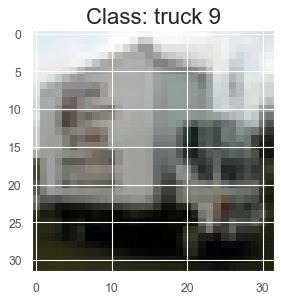}
 \vspace{1mm}
\end{subfigure}
\begin{subfigure}[t]{.25\textwidth}
 \centering
 \includegraphics[width=1\linewidth]{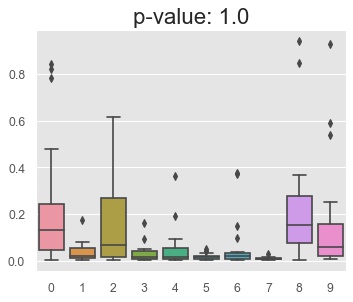}
 \vspace{-3mm}
\end{subfigure}
\begin{subfigure}[t]{.25\textwidth}
 \centering
 \includegraphics[width=1\linewidth]{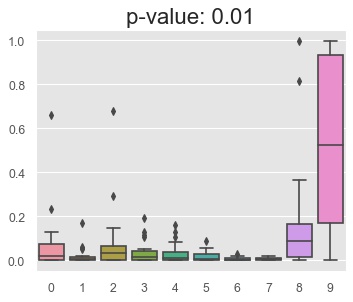}
 \vspace{-3mm}
\end{subfigure}
\begin{subfigure}[t]{.25\textwidth}
 \centering
 \includegraphics[width=1\linewidth]{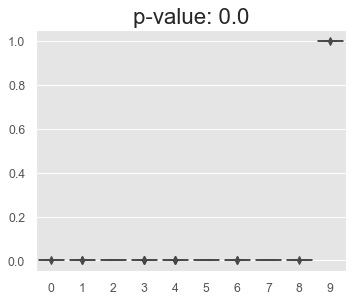}
 \vspace{-3mm}

\end{subfigure}%

\begin{subfigure}[t]{.25\textwidth}
 \centering
 \includegraphics[width=0.8\linewidth]{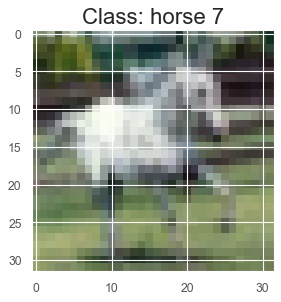}\par
 \vspace{1mm}
\end{subfigure}
\begin{subfigure}[t]{.25\textwidth}
 \centering
 \includegraphics[width=1\linewidth]{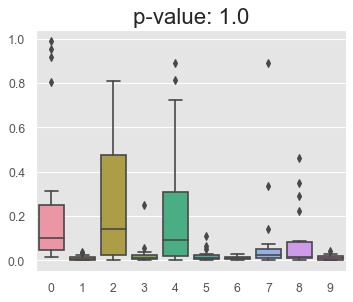}\par
 \vspace{-3mm}
\end{subfigure}
\begin{subfigure}[t]{.25\textwidth}
 \centering
 \includegraphics[width=1\linewidth]{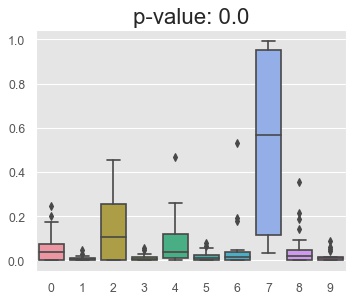}
 \vspace{-3mm}
\end{subfigure}
\begin{subfigure}[t]{.25\textwidth}
 \centering
 \includegraphics[width=1\linewidth]{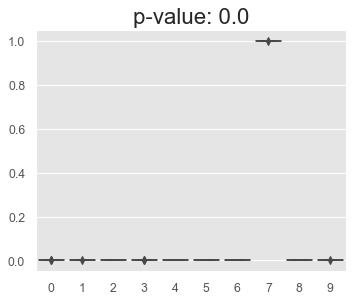}
 \vspace{-3mm}

\end{subfigure}%

\begin{subfigure}[t]{.25\textwidth}
 \centering
 \includegraphics[width=0.8\linewidth]{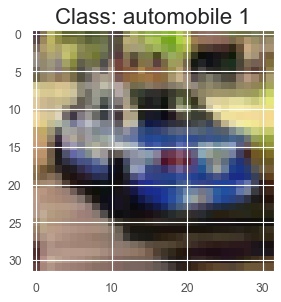}\par
 \vspace{1mm}
\end{subfigure}
\begin{subfigure}[t]{.25\textwidth}
 \centering
 \includegraphics[width=1\linewidth]{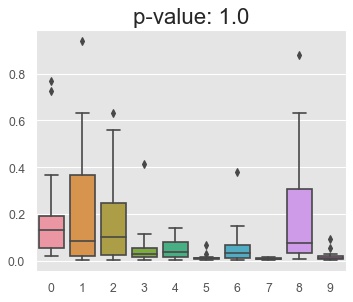}\par
 \vspace{-3mm}
\end{subfigure}
\begin{subfigure}[t]{.25\textwidth}
 \centering
 \includegraphics[width=1\linewidth]{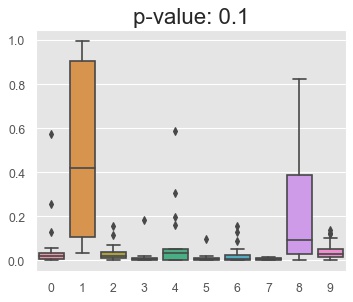}
 \vspace{-3mm}
\end{subfigure}
\begin{subfigure}[t]{.25\textwidth}
 \centering
 \includegraphics[width=1\linewidth]{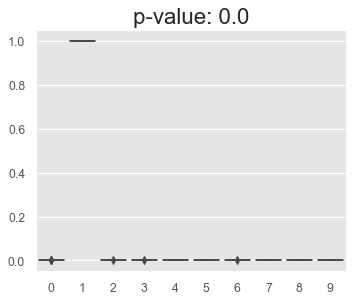}
 \vspace{-3mm}

\end{subfigure}%

\begin{subfigure}[t]{.25\textwidth}
 \centering
 \includegraphics[width=0.8\linewidth]{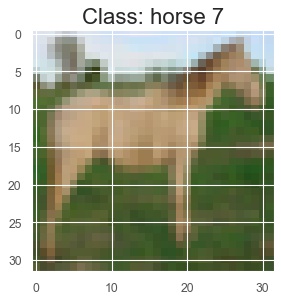}\par
 \vspace{1mm}
\end{subfigure}
\begin{subfigure}[t]{.25\textwidth}
 \centering
 \includegraphics[width=1\linewidth]{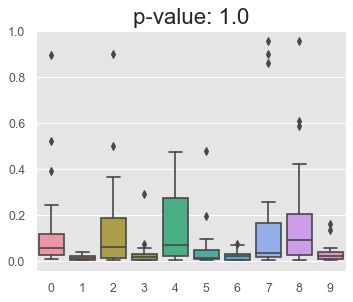}\par
 \vspace{-3mm}
\end{subfigure}
\begin{subfigure}[t]{.25\textwidth}
 \centering
 \includegraphics[width=1\linewidth]{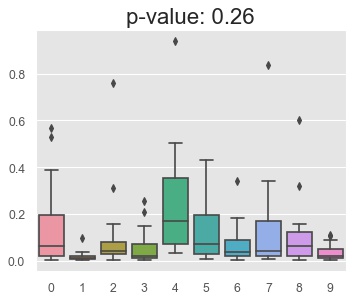}
 \vspace{-3mm}
\end{subfigure}
\begin{subfigure}[t]{.25\textwidth}
 \centering
 \includegraphics[width=1\linewidth]{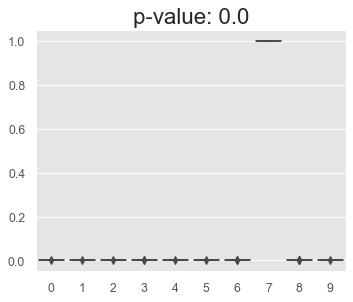}
 \vspace{-3mm}

\end{subfigure}%

\begin{subfigure}[t]{.25\textwidth}
 \centering
 \includegraphics[width=0.8\linewidth]{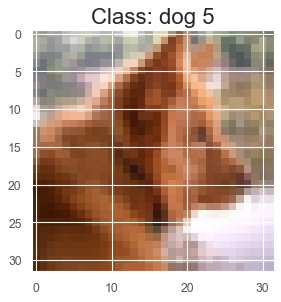}
 \caption{  Image}
 \vspace{1mm}
\end{subfigure}
\begin{subfigure}[t]{.25\textwidth}
 \centering
 \includegraphics[width=1\linewidth]{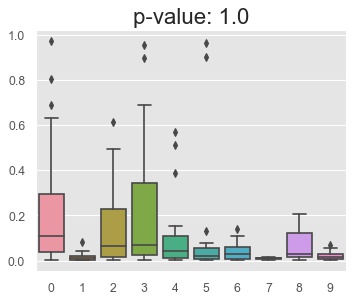}
 \caption{  Concrete Dropout}
 \vspace{-3mm}
\end{subfigure}
\begin{subfigure}[t]{.25\textwidth}
 \centering
 \includegraphics[width=1\linewidth]{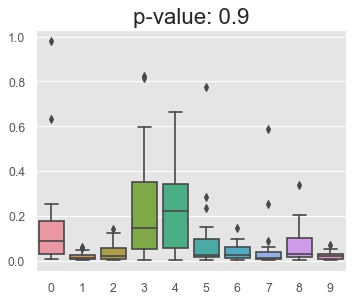}
 \caption{  MC Dropout}
 \vspace{-3mm}
\end{subfigure}
\begin{subfigure}[t]{.25\textwidth}
 \centering
 \includegraphics[width=1\linewidth]{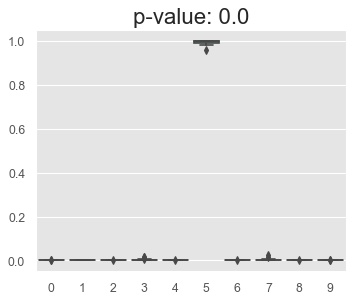}
 \caption{  Contextual Dropout}
 \vspace{-3mm}

\end{subfigure}%
\caption{  Visualization of probability outputs of different dropouts on CIFAR-10. 5 plots that {\bf{Concrete Dropout}} is the most uncertain are presented. Number to class map: \{0: airplane, 1: automobile, 2: bird, 3: cat, 4: deer, 5: dog, 6: frog, 7: horse, 8: ship, 9: truck.\}}
\label{fig:CIFAR-10_concrete}\vspace{-3mm}
\end{figure*}

\begin{figure*}[h]

\begin{subfigure}[t]{.25\textwidth}
 \centering
 \includegraphics[width=0.8\linewidth]{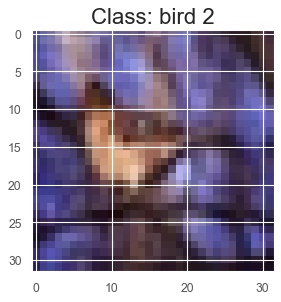}
 \vspace{1mm}
\end{subfigure}
\begin{subfigure}[t]{.25\textwidth}
 \centering
 \includegraphics[width=1\linewidth]{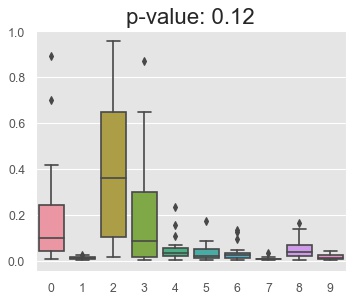}
 \vspace{-3mm}
\end{subfigure}
\begin{subfigure}[t]{.25\textwidth}
 \centering
 \includegraphics[width=1\linewidth]{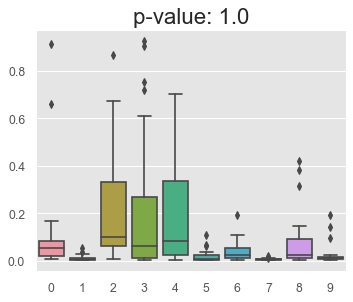}
 \vspace{-3mm}
\end{subfigure}
\begin{subfigure}[t]{.25\textwidth}
 \centering
 \includegraphics[width=1\linewidth]{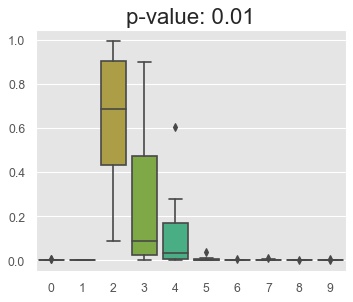}
 \vspace{-3mm}

\end{subfigure}%

\begin{subfigure}[t]{.25\textwidth}
 \centering
 \includegraphics[width=0.8\linewidth]{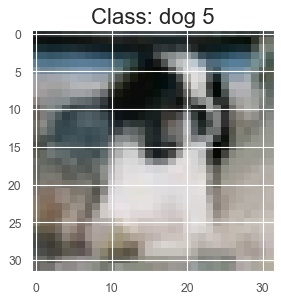}\par
 \vspace{1mm}
\end{subfigure}
\begin{subfigure}[t]{.25\textwidth}
 \centering
 \includegraphics[width=1\linewidth]{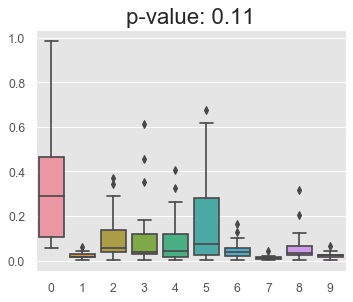}\par
 \vspace{-3mm}
\end{subfigure}
\begin{subfigure}[t]{.25\textwidth}
 \centering
 \includegraphics[width=1\linewidth]{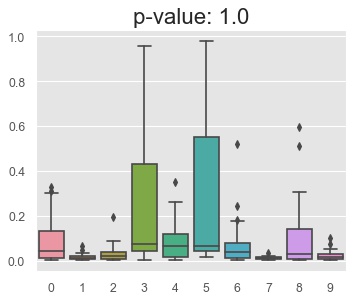}
 \vspace{-3mm}
\end{subfigure}
\begin{subfigure}[t]{.25\textwidth}
 \centering
 \includegraphics[width=1\linewidth]{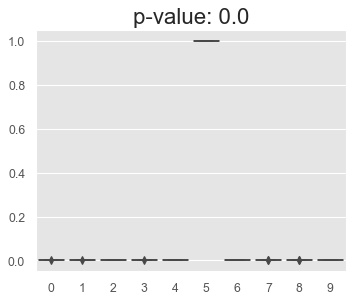}
 \vspace{-3mm}

\end{subfigure}%

\begin{subfigure}[t]{.25\textwidth}
 \centering
 \includegraphics[width=0.8\linewidth]{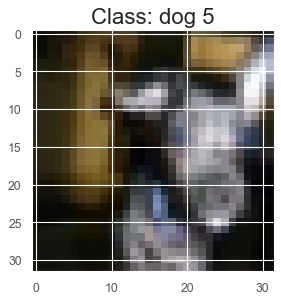}\par
 \vspace{1mm}
\end{subfigure}
\begin{subfigure}[t]{.25\textwidth}
 \centering
 \includegraphics[width=1\linewidth]{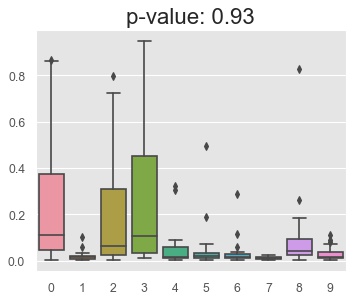}\par
 \vspace{-3mm}
\end{subfigure}
\begin{subfigure}[t]{.25\textwidth}
 \centering
 \includegraphics[width=1\linewidth]{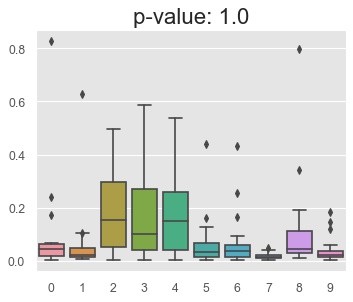}
 \vspace{-3mm}
\end{subfigure}
\begin{subfigure}[t]{.25\textwidth}
 \centering
 \includegraphics[width=1\linewidth]{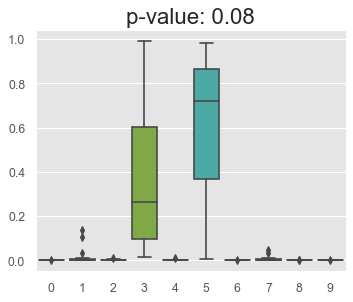}
 \vspace{-3mm}

\end{subfigure}%

\begin{subfigure}[t]{.25\textwidth}
 \centering
 \includegraphics[width=0.8\linewidth]{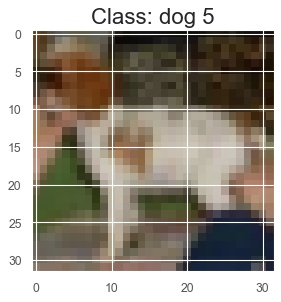}\par
 \vspace{1mm}
\end{subfigure}
\begin{subfigure}[t]{.25\textwidth}
 \centering
 \includegraphics[width=1\linewidth]{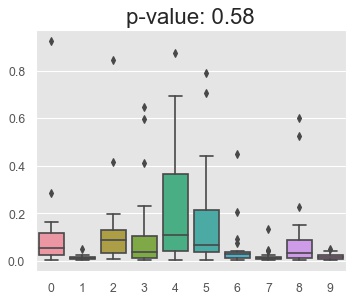}\par
 \vspace{-3mm}
\end{subfigure}
\begin{subfigure}[t]{.25\textwidth}
 \centering
 \includegraphics[width=1\linewidth]{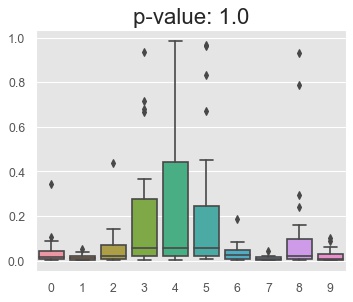}
 \vspace{-3mm}
\end{subfigure}
\begin{subfigure}[t]{.25\textwidth}
 \centering
 \includegraphics[width=1\linewidth]{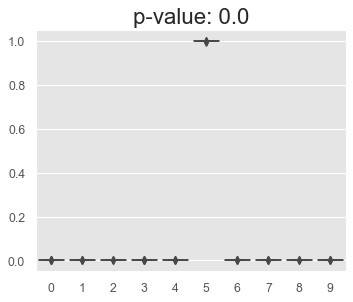}
 \vspace{-3mm}

\end{subfigure}%

\begin{subfigure}[t]{.25\textwidth}
 \centering
 \includegraphics[width=0.8\linewidth]{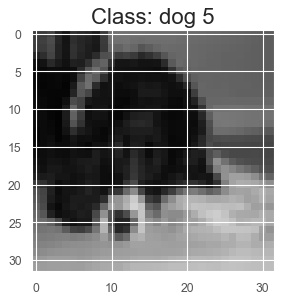}
 \caption{  Image}
 \vspace{1mm}
\end{subfigure}
\begin{subfigure}[t]{.25\textwidth}
 \centering
 \includegraphics[width=1\linewidth]{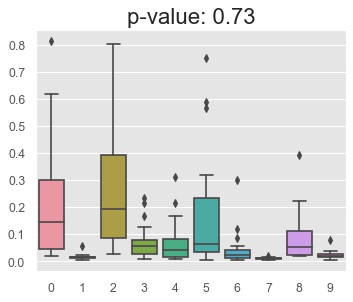}
 \caption{  Concrete Dropout}
 \vspace{-3mm}
\end{subfigure}
\begin{subfigure}[t]{.25\textwidth}
 \centering
 \includegraphics[width=1\linewidth]{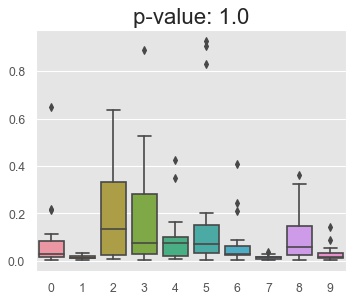}
 \caption{  MC Dropout}
 \vspace{-3mm}
\end{subfigure}
\begin{subfigure}[t]{.25\textwidth}
 \centering
 \includegraphics[width=1\linewidth]{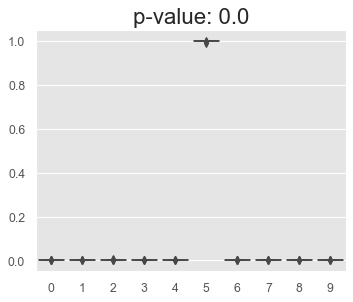}
 \caption{  Contextual Dropout}
 \vspace{-3mm}

\end{subfigure}%
\caption{  Visualization of probability outputs of different dropouts on CIFAR-10. 5 plots that {\bf{MC Dropout}} is the most uncertain are presented. Number to class map: \{0: airplane, 1: automobile, 2: bird, 3: cat, 4: deer, 5: dog, 6: frog, 7: horse, 8: ship, 9: truck.\}}
\label{fig:CIFAR-10_mc}\vspace{-3mm}
\end{figure*}

\begin{figure}[!htb]

\begin{subfigure}[t]{.25\textwidth}
 \centering
 \includegraphics[width=0.8\linewidth]{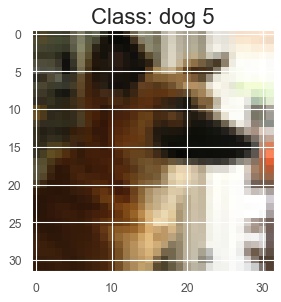}
 \vspace{1mm}
\end{subfigure}
\begin{subfigure}[t]{.25\textwidth}
 \centering
 \includegraphics[width=1\linewidth]{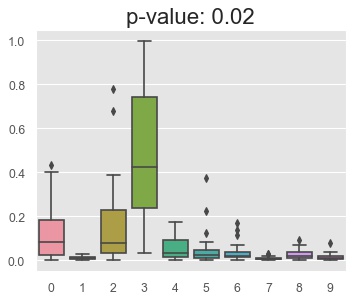}
 \vspace{-3mm}
\end{subfigure}
\begin{subfigure}[t]{.25\textwidth}
 \centering
 \includegraphics[width=1\linewidth]{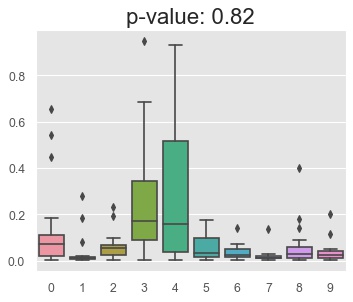}
 \vspace{-3mm}
\end{subfigure}
\begin{subfigure}[t]{.25\textwidth}
 \centering
 \includegraphics[width=1\linewidth]{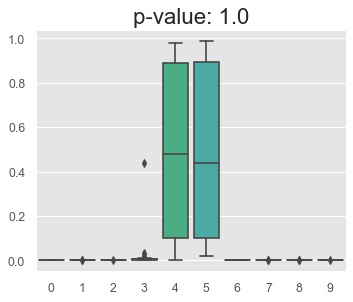}
 \vspace{-3mm}

\end{subfigure}%

\begin{subfigure}[t]{.25\textwidth}
 \centering
 \includegraphics[width=0.8\linewidth]{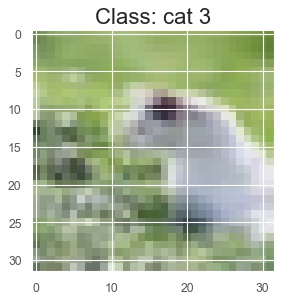}\par
 \vspace{1mm}
\end{subfigure}
\begin{subfigure}[t]{.25\textwidth}
 \centering
 \includegraphics[width=1\linewidth]{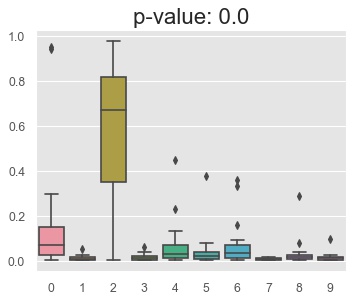}\par
 \vspace{-3mm}
\end{subfigure}
\begin{subfigure}[t]{.25\textwidth}
 \centering
 \includegraphics[width=1\linewidth]{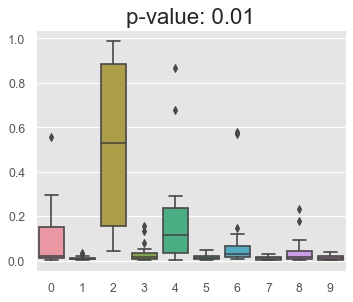}
 \vspace{-3mm}
\end{subfigure}
\begin{subfigure}[t]{.25\textwidth}
 \centering
 \includegraphics[width=1\linewidth]{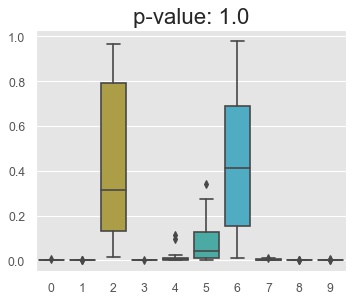}
 \vspace{-3mm}

\end{subfigure}%

\begin{subfigure}[t]{.25\textwidth}
 \centering
 \includegraphics[width=0.8\linewidth]{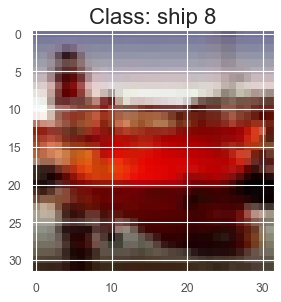}\par
 \vspace{1mm}
\end{subfigure}
\begin{subfigure}[t]{.25\textwidth}
 \centering
 \includegraphics[width=1\linewidth]{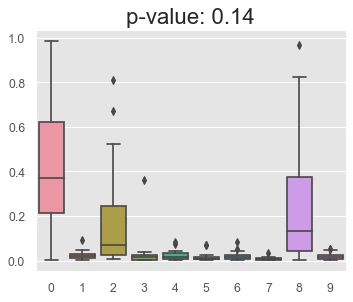}\par
 \vspace{-3mm}
\end{subfigure}
\begin{subfigure}[t]{.25\textwidth}
 \centering
 \includegraphics[width=1\linewidth]{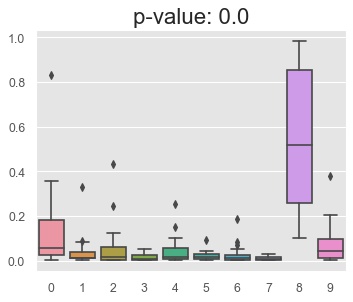}
 \vspace{-3mm}
\end{subfigure}
\begin{subfigure}[t]{.25\textwidth}
 \centering
 \includegraphics[width=1\linewidth]{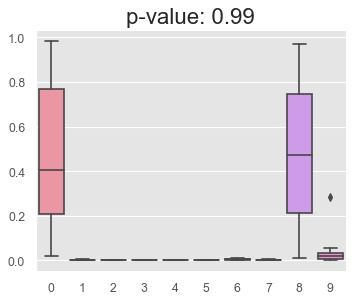}
 \vspace{-3mm}

\end{subfigure}%

\begin{subfigure}[t]{.25\textwidth}
 \centering
 \includegraphics[width=0.8\linewidth]{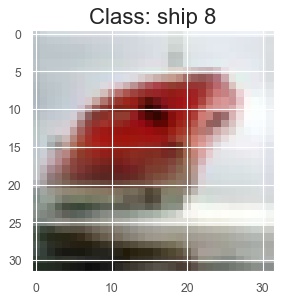}\par
 \vspace{1mm}
\end{subfigure}
\begin{subfigure}[t]{.25\textwidth}
 \centering
 \includegraphics[width=1\linewidth]{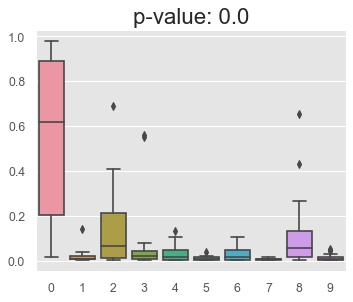}\par
 \vspace{-3mm}
\end{subfigure}
\begin{subfigure}[t]{.25\textwidth}
 \centering
 \includegraphics[width=1\linewidth]{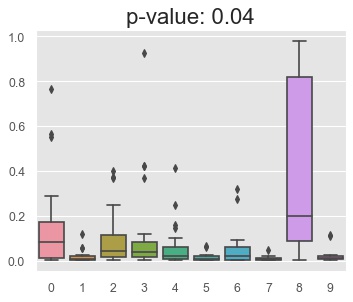}
 \vspace{-3mm}
\end{subfigure}
\begin{subfigure}[t]{.25\textwidth}
 \centering
 \includegraphics[width=1\linewidth]{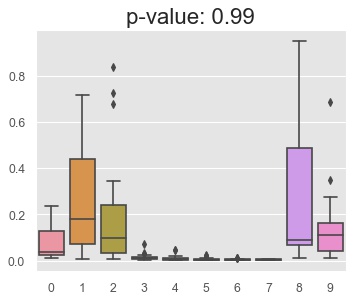}
 \vspace{-3mm}

\end{subfigure}%

\begin{subfigure}[t]{.25\textwidth}
 \centering
 \includegraphics[width=0.8\linewidth]{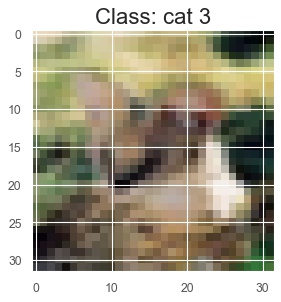}
 \caption{  Image}
 \vspace{1mm}
\end{subfigure}
\begin{subfigure}[t]{.25\textwidth}
 \centering
 \includegraphics[width=1\linewidth]{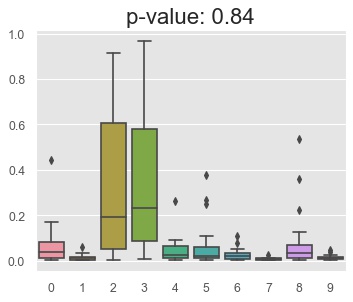}
 \caption{  Concrete Dropout}
 \vspace{-3mm}
\end{subfigure}
\begin{subfigure}[t]{.25\textwidth}
 \centering
 \includegraphics[width=1\linewidth]{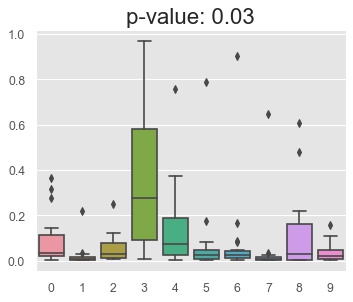}
 \caption{  MC Dropout}
 \vspace{-3mm}
\end{subfigure}
\begin{subfigure}[t]{.25\textwidth}
 \centering
 \includegraphics[width=1\linewidth]{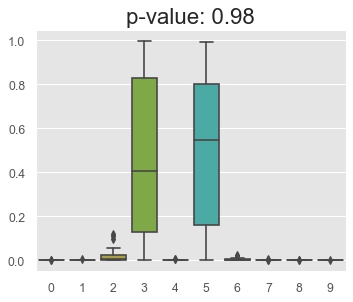}
 \caption{  Contextual Dropout}
 \vspace{-3mm}

\end{subfigure}%
\caption{  Visualization of probability outputs of different dropouts on CIFAR-10. 5 plots that {\bf Contextual Dropout} is the most uncertain are presented. Number to class map: \{0: airplane, 1: automobile, 2: bird, 3: cat, 4: deer, 5: dog, 6: frog, 7: horse, 8: ship, 9: truck.\}}
\label{fig:CIFAR-10_rbnn}\vspace{-3mm}
\end{figure}

In this section, we visualize $5$ most uncertain images for each dropout (only include Bernoulli, Concrete, and Contextual Bernoulli dropout for simplicity) leading to $15$ images in total. 
The true images with the labels are on the left side and boxplots of probability distributions of different dropouts are on the right side. All models are trained on the original CIFAR-10 dataset. Among these $15$ images, we observe that contextual dropout predicts the right answer if it is certain, and it is certain and predicts the right answer on many images that MC dropout or concrete dropout is uncertain about (e.g, many images in Figure~\ref{fig:CIFAR-10_concrete}-\ref{fig:CIFAR-10_mc}). However, MC dropout or concrete dropout is uncertain about some easy examples (images in Figures~\ref{fig:CIFAR-10_concrete}-\ref{fig:CIFAR-10_mc}) or certain on some wrong predictions (images in Figure~\ref{fig:CIFAR-10_rbnn}).  Moreover, on an image that all three methods have high uncertainty, concrete dropout often places a higher probability on the correct answer than the other two methods (images in Figure~\ref{fig:CIFAR-10_rbnn}).



\clearpage
\subsection{Visualization for Visual Question Answering}
\label{sec:vqa_visualization}
\begin{figure}[!htp]
 \centering
 \includegraphics[width=0.9\textwidth,height=8.0cm]{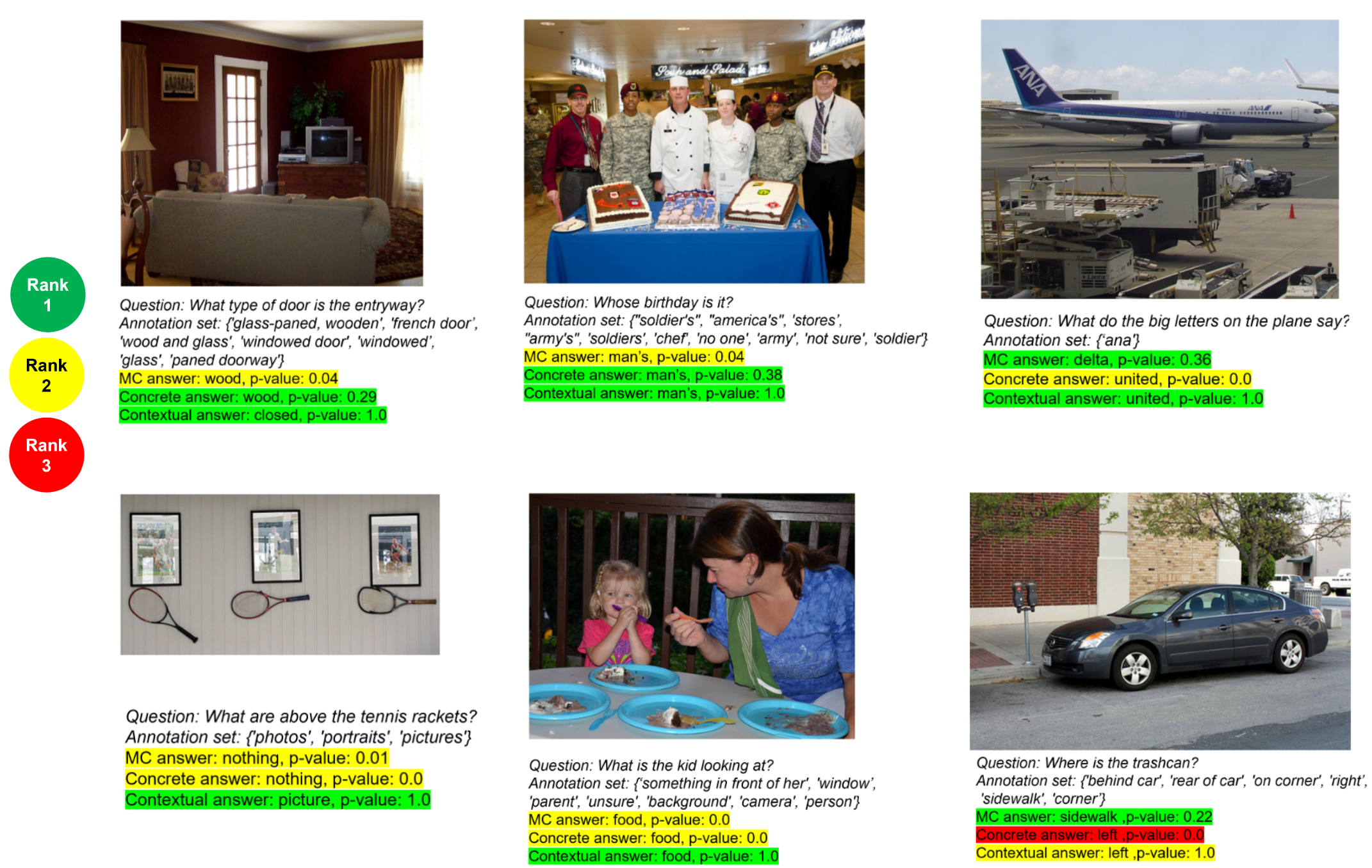}
 \vspace{-1.5mm}
 \caption{  VQA visualization: $6$ plots that {\bf Contextual Dropout} is the most uncertain are presented.}
 \label{fig:vqa_vis_contextual} \vspace{3mm}
\end{figure}

\begin{figure}[!htp]
 \centering
 \includegraphics[width=0.9\textwidth,height=8.0cm]{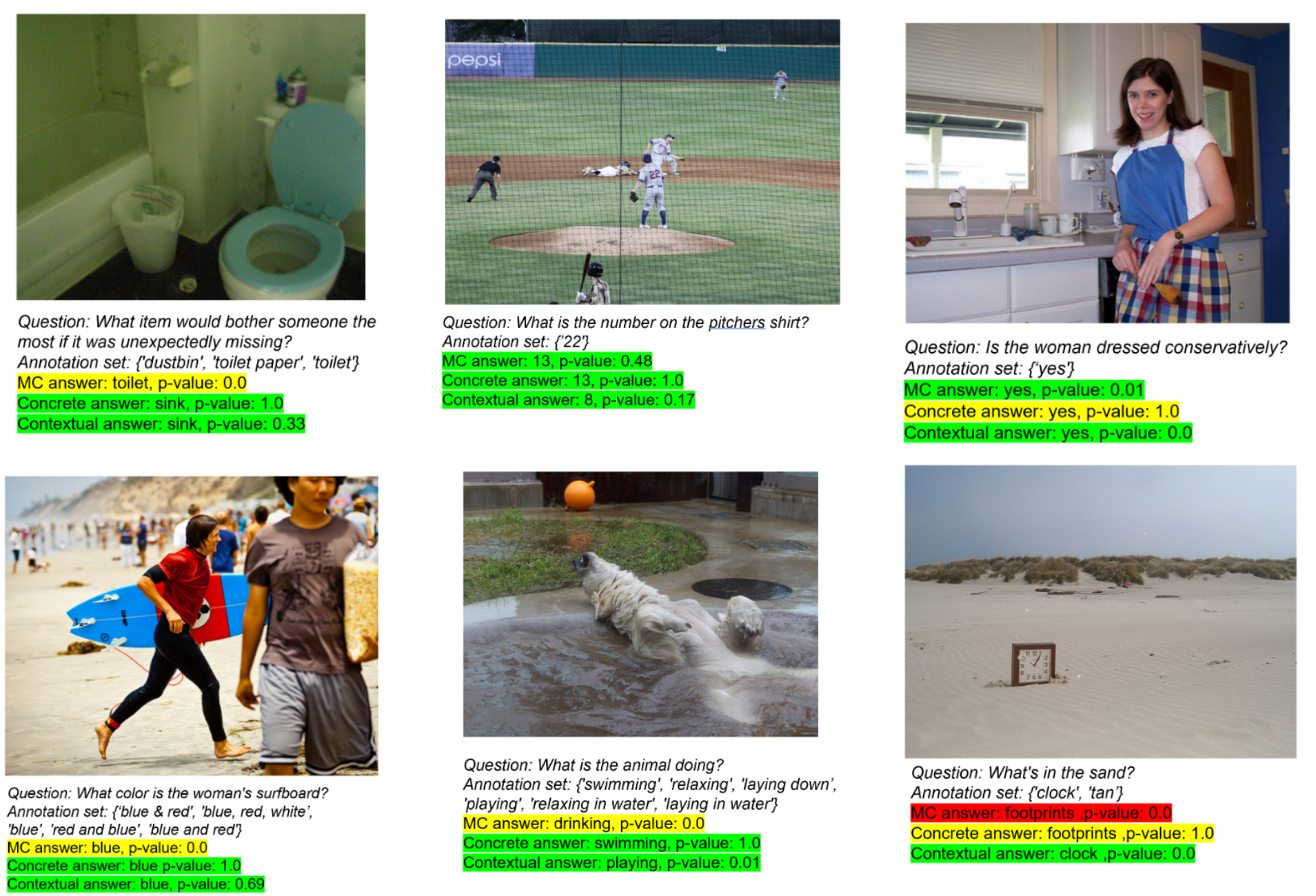}
 \vspace{-1.5mm}
 \caption{  VQA visualization: $6$ plots that {\bf Concrete Dropout} is the most uncertain are presented.}
 \label{fig:vqa_vis_concrete} \vspace{-3mm}
\end{figure}

\begin{figure}[t]
 \centering
 \includegraphics[width=0.9\textwidth,height=8.0cm]{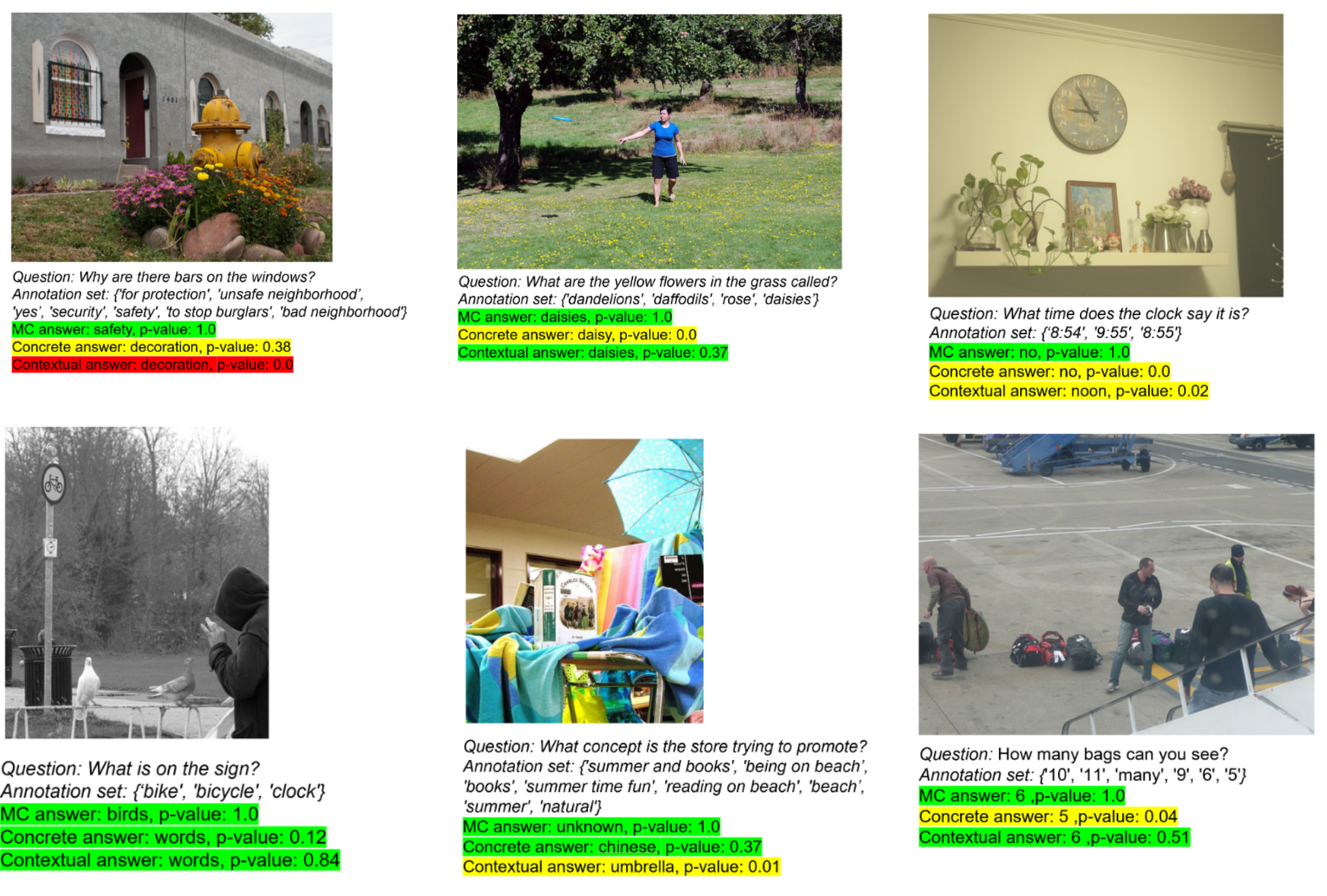}
 \vspace{-1.5mm}
 \caption{  VQA visualization: $6$ plots that {\bf MC Dropout} is the most uncertain are presented.}
 \label{fig:vqa_vis_mc} \vspace{-3mm}
\end{figure}

\begin{figure}[!htp]
 \centering
 \includegraphics[width=0.9\textwidth,height=8.0cm]{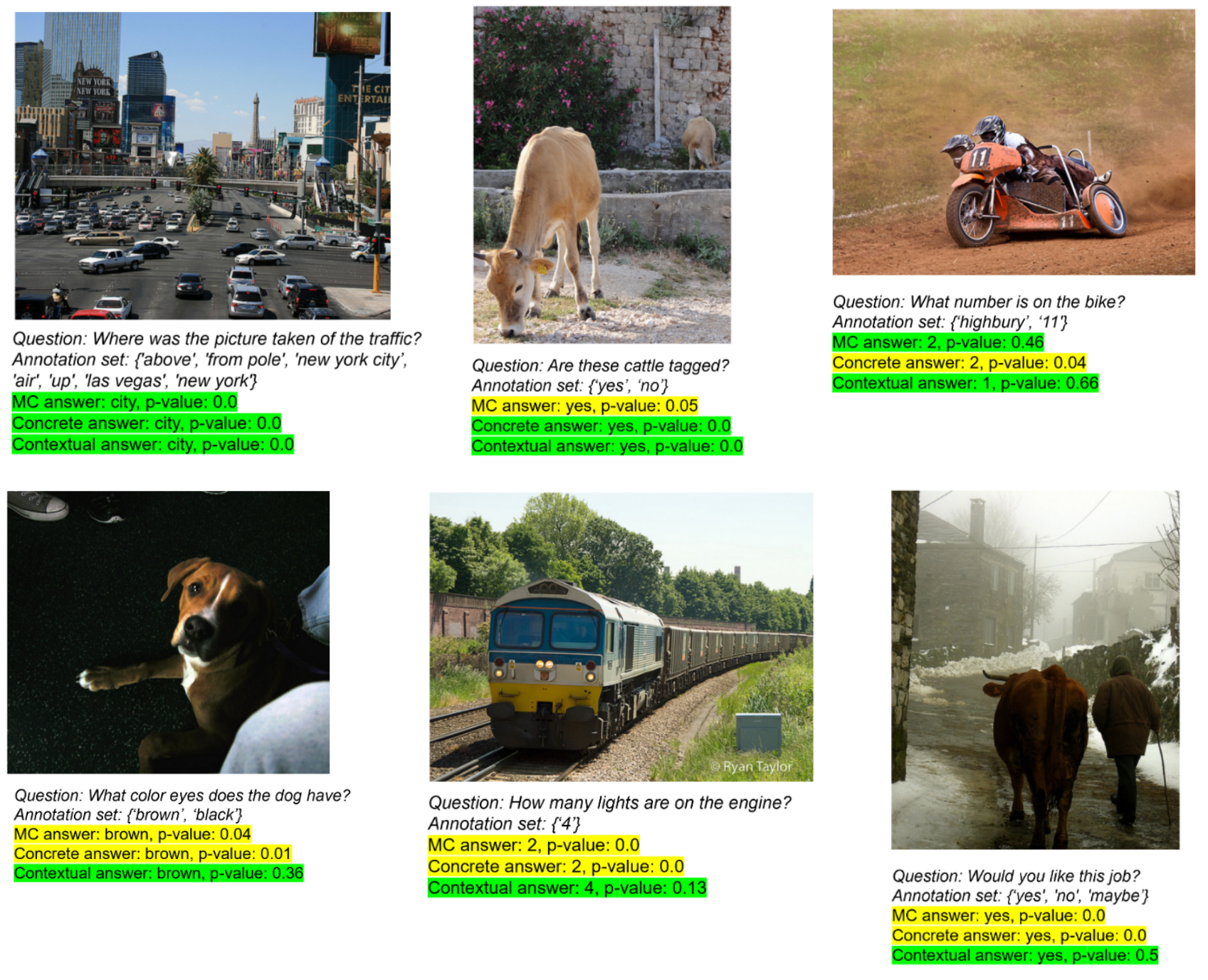}
 \includegraphics[width=0.9\textwidth,height=8.0cm]{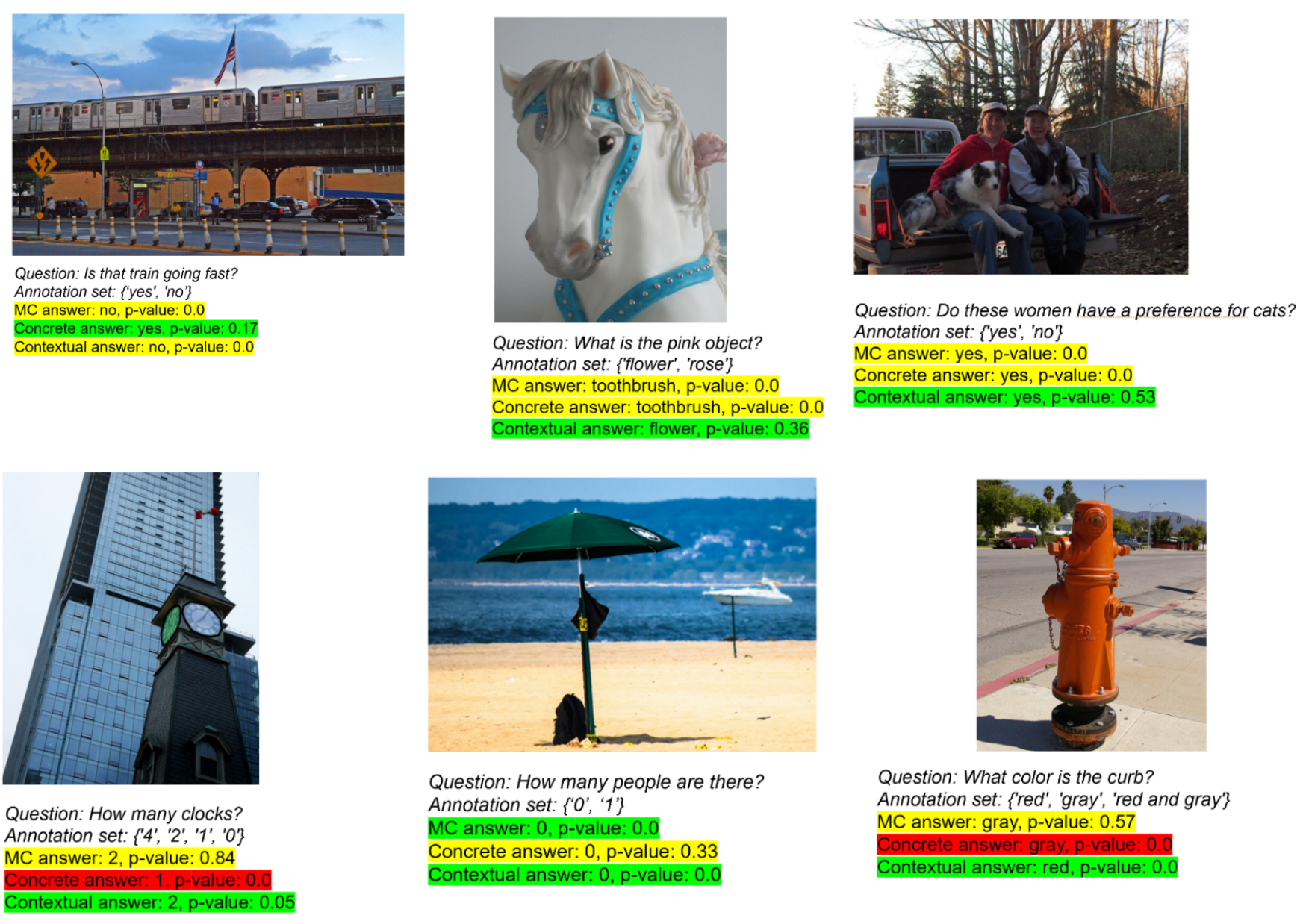}
 \vspace{-1.5mm}
 \caption{  VQA visualization: $12$ randomly selected plots are presented.}
 \label{fig:vqa_vis_random} \vspace{3mm}
\end{figure}

In Figures~\ref{fig:vqa_vis_contextual}-\ref{fig:vqa_vis_random}, we visualize some image-question pairs, along with the human annotations (for simplicity, we only show the different answers in the annotation set) and compare the predictions and uncertainty estimations of different dropouts (only include Bernoulli dropout, Concrete dropout, and contextual Bernoulli dropout) on the noisy data. We include $12$ randomly selected image-question pairs, and $6$ most uncertain image-question pairs for each dropout as challenging samples ($30$ in total). For each sample, we manually rank different methods by the general rule that accurate and certain is the most preferred, followed by accurate and uncertain, inaccurate and uncertain, and then inaccurate and certain. 
For each image-question pair, we rank three different dropouts based on their answers and $p$-values, and highlight the best performing one, the second best, and the worst with green, yellow, and red, respectively (tied ranks are allowed).  
As shown in the plots, overall contextual dropout is more conservative on its wrong predictions and more certain on its correct predictions than other methods for both randomly selected images and challenging images.



\end{document}